\documentclass[journal]{IEEEtran}

\ifCLASSINFOpdf
\else
\fi
\usepackage{makecell}
\usepackage{graphicx}
\usepackage{cite}
\usepackage{amsmath,amssymb,amsfonts,bbding}
\usepackage{algorithmic}
\usepackage{textcomp}
\usepackage[caption=false,font=footnotesize]{subfig}
\usepackage{algorithm}
\usepackage{amsmath,bm}
\usepackage{multirow}
\usepackage{threeparttable}
\usepackage{color}
\usepackage{colortbl}
\usepackage{xcolor}
\usepackage{hyperref}
\usepackage{stfloats}
\usepackage[sort&compress,numbers]{natbib}
\usepackage{comment}

\begin{document}

%
\title{Parameter-Efficient Transfer Learning for \\ Remote Sensing Image-Text Retrieval}

\author{Yuan Yuan,~\IEEEmembership{Senior Member, IEEE}, Yang Zhan, and~Zhitong Xiong,~\IEEEmembership{Member, IEEE}

\thanks{
This work was supported in part by grants from the Innovation Foundation for Doctor Dissertation of Northwestern Polytechnical University (No.CX2023030), the National Key R\&D Program of China (No.2020YFB2103900), and the National Science Fund for Distinguished Young Scholars (No.61825603). \textit{(Corresponding authors: Yuan Yuan and Zhitong Xiong.)}}
\thanks{
Yang Zhan and Yuan Yuan are with the School of Artificial Intelligence, Optics, and Electronics (iOPEN), Northwestern Polytechnical University, Xi'an 710072, China (e-mail:\href{mailto:zhanyang@mail.nwpu.edu.cn}{zhanyang@mail.nwpu.edu.cn}; \href{mailto:y.yuan@nwpu.edu.cn}{y.yuan@nwpu.edu.cn}).}
\thanks{Zhitong Xiong is with the Chair of Data Science in Earth Observation, Technical University of Munich (TUM), 80333 Munich, Germany.}
}


\definecolor{XZT}{RGB}{0,0,255}
\definecolor{ZY}{RGB}{255,0,0}

\maketitle

\begin{abstract}
    Vision-and-language pre-training (VLP) models have experienced a surge in popularity recently. By fine-tuning them on specific datasets, significant performance improvements have been observed in various tasks. 
    However, full fine-tuning of VLP models not only consumes a significant amount of computational resources but also has a significant environmental impact. Moreover, as remote sensing (RS) data is constantly being updated, full fine-tuning may not be practical for real-world applications. To address this issue, in this work, we investigate the parameter-efficient transfer learning (PETL) method to effectively and efficiently transfer visual-language knowledge from the natural domain to the RS domain on the image-text retrieval task. To this end, we make the following contributions. 1) We construct a novel and sophisticated PETL framework for the RS image-text retrieval (RSITR) task, which includes the pretrained CLIP model, a multimodal remote sensing adapter, and a hybrid multi-modal contrastive (HMMC) learning objective; 2) To deal with the problem of high intra-modal similarity in RS data, we design a simple yet effective HMMC loss; 
    3) We provide comprehensive empirical studies for PETL-based RS image-text retrieval. Our results demonstrate that the proposed method is promising and of great potential for practical applications.
    4) We benchmark extensive state-of-the-art PETL methods on the RSITR task. Our proposed model only contains 0.16M training parameters, which can achieve a parameter reduction of 98.9\% compared to full fine-tuning, resulting in substantial savings in training costs. Our retrieval performance exceeds traditional methods by 7-13\% and achieves comparable or better performance than full fine-tuning. This work can provide new ideas and useful insights for RS vision-language tasks.
\end{abstract}

\begin{IEEEkeywords}
Parameter-Efficient Transfer Learning (PETL), adapter, cross-modal, remote sensing image-text retrieval.
\end{IEEEkeywords}

\IEEEpeerreviewmaketitle

\section{Introduction}
\label{sec:introduction}
\IEEEPARstart{W}{ith} the development of Earth observation technology \cite{earthnets4eo}, 
remote sensing (RS) imagery is becoming more and more accessible, improving human's perception of the Earth \cite{9775021, 10005113}. However, how to efficiently convert RS imagery into actionable information is still significant research \cite{9524842,9760473,9511333,9795332}.
In order to fully exploit the potential of RS images in human-computer interaction, RS vision-language (VL) tasks have become a hot research topic in recent years. The different granularity of VL multi-modal tasks have been introduced into RS data, including image level \cite{9745546,9594840,9437331,8240966}, object level \cite{yuan2022easy,10056343, 3503161.3548316}, pixel level \cite{9893840}, and spatial-temporal level \cite{9934924,yuan2022change}. 
These technologies have promising applications in urban planning, disaster monitoring, search and rescue activities, resource detection, and agricultural production \cite{9371014,9088993,10093906,9868210,10050427,9540028}.

\begin{figure}[t]	
	\centering
	\includegraphics[width=1\linewidth]{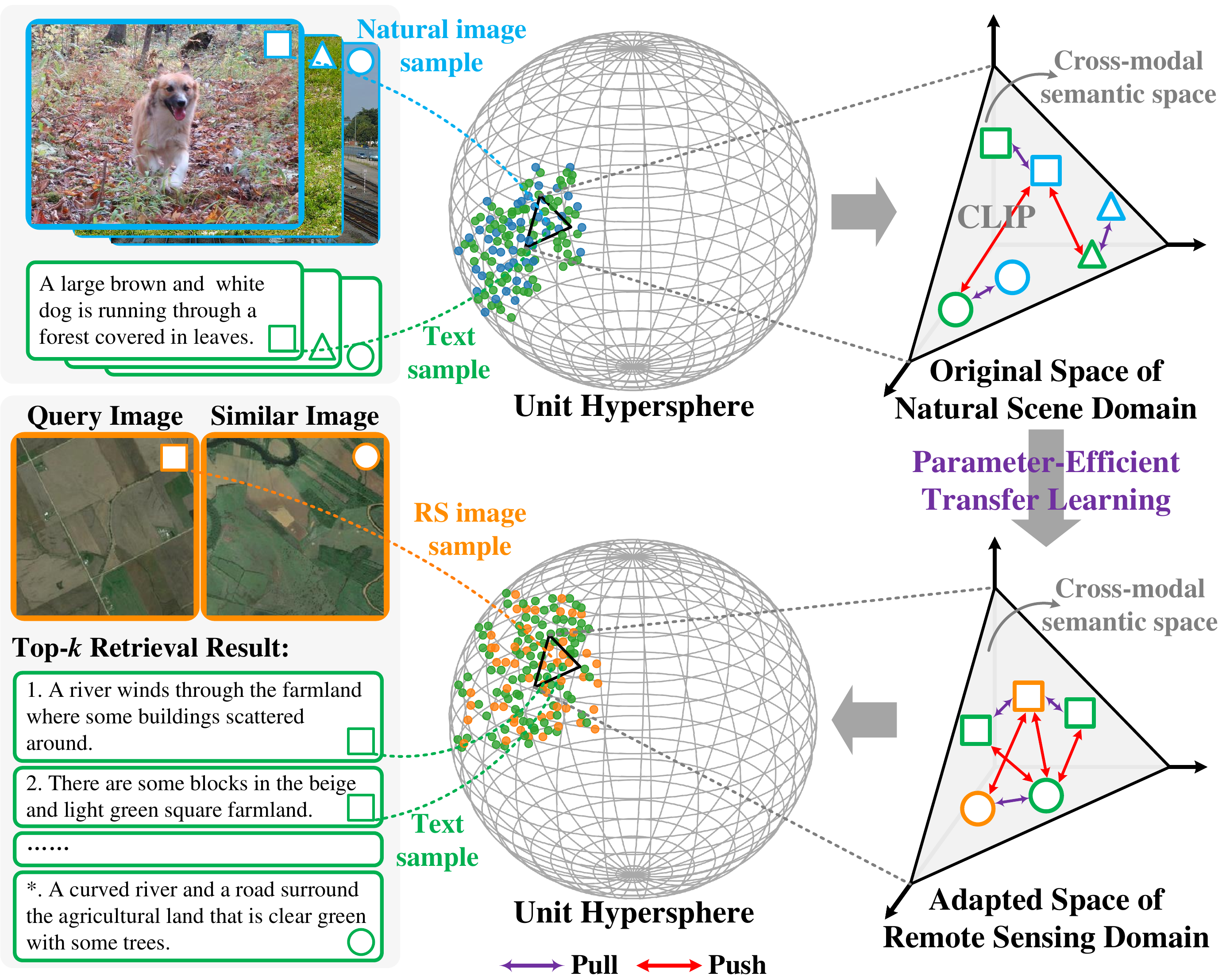}
	\caption{The matched natural image-text samples have the same vector direction in the unit hypersphere space of the pre-trained CLIP model. The PETL method learns specific knowledge of RS domain to get the adapted space. In the original space of the natural scene domain, only the distance between different modalities is paid attention to. However, in the RS domain, it is necessary to consider samples with high intra-modal similarity to avoid the problem of matching errors.
    }
	\label{fig:RSPETL}
\end{figure}

Large vision-and-language pre-training (VLP) models have surged \cite{chen2023vlp, du2022survey} in recent years. In particular, contrastive vision-language pre-training (CLIP) \cite{pmlr-v139-radford21a} has shown great potential in multi-modal representations and can project natural image and text modalities into a joint semantic subspace. As shown in Fig. \ref{fig:RSPETL}, the aligned image-text samples have the same vector direction on the unit hypersphere.
Fine-tuning large VLP models has become a fundamental paradigm of research. However, the research of transferring the knowledge learned from image-text pairs in the natural domain to a more complex RS domain is still under-explored. 

Meanwhile, VLP models in the RS domain have not been proposed due to the challenges of RS vision-language tasks. Although there are a large number of publicly available RS images, few of them are captioned and even fewer are multi-captioned.
The fully fine-tuned CLIP model has achieved encouraging performance for RS image classification$\footnote{\label{note1}https://github.com/arampacha/CLIP-rsicd}$.
However, this approach is not feasible due to the heavy computation, memory storage, and excessive $CO_{2}$ emissions. As RS images are updated, it is impractical to adapt with constant full fine-tuning on a daily basis. In this context, we explore a new research paradigm of parameter-efficient transfer learning (PETL) from the natural scene domain to the RS domain. Due to the outstanding representational capability of CLIP in VL, we use the CLIP model to study RS image-text retrieval (RSITR). This leads to a new research task, namely PETL-based RS image-text retrieval (PE-RSITR). The RSITR task can verify the performance of the adapted space of the RS domain, as shown in Fig. \ref{fig:RSPETL}. 

Nowadays, the mainstream methods of PETL are mainly divided into adapter \cite{houlsby19a} and prompt learning \cite{li2021prefix, lester2021power}. PETL only fine-tunes a small number of parameters while keeping the parameters of the CLIP unchanged, which greatly reduces the computational cost while having comparable performance to full fine-tuning.
However, existing works typically focus on downstream tasks from the same domain of the VLP models. This creates a limit that a strong VLP model with sufficient knowledge may not be available in an unknown specific domain (\textit{e.g.}, remote sensing). Therefore, there are still many challenges to exploring PE-RSITR. 

First, CLIP is pre-trained in the natural scene domain, which has domain gaps with the RS domain. To bridge the significant domain gap, the knowledge of CLIP needs to be transferred from "VL of the natural scene" to "VL of the RS". John von Neumann once said: with four parameters I can fit an elephant, and with five I can make him wiggle his trunk. Therefore, we attempt to design a method with a small number of trained parameters to explore the new knowledge of RS image-text efficiently while inheriting the prior knowledge structure of the natural scene domain appropriately. 
In addition, the RSITR task involves two modalities. If there is no cross-modal interaction mechanism, only suboptimal results can be obtained \cite{Jiang10.48550,2302.06605}. Therefore, we further try to design a method that does not increase parameters and can accomplish cross-modal knowledge sharing.
Finally, the RS image-text data is very different from the data of the natural domain. The RS images are collected by satellites from an overhead view and the intra-class similarity is extremely high due to the earth's texture. The visualization results of the textual similarity in the literature \cite{9437331} show that the caption similarity is also high. Since this is not fully considered by existing methods, RSITR often results in the error of misalignment of similar RS images or captions.

To tackle the above problems, we propose a novel and sophisticated PE-RSITR framework. Although adapter-based methods have been widely explored in prior works, it is still non-trivial to design an effective adapter for the RSITR task. Based on extensive experiments and explorations, we make the following design decisions. 
1) We design a more compact multimodal remote sensing adapter (MRS-Adapter) that has no skip connection and connects only once in parallel with the transformer block. 2) Inspired by the Cross-Modal Adapter \cite{Jiang10.48550}, MRS-Adapter utilizes a linear layer for weight sharing. The shared linear layer enables the fine-grained information of RS image modality and text modality to interact, which can enhance the RS vision language modality representation.

Furthermore, token-level data augmentation is designed to construct intra-modal positive pairs for RS images and texts. The method of data augmentation is to adopt the simple random dropout. By devising a simple yet efficient loss function of hybrid multi-modal contrastive constraints without increasing parameters, the distance between the query image (query text) and other similar images (similar texts) can be pushed to avoid matching errors.
We have conducted extensive experiments on three commonly used datasets, \textit{i.e.}, RSICD \cite{8240966}, RSITMD \cite{9437331}, and UCM \cite{7546397}. First, we benchmark many state-of-the-art (SOTA) methods on the PE-RSITR task, and the adapter largely outperforms the prompt learning method.
Secondly, the trained parameter of our proposed method is 0.16M, which can reduce 98.9\% parameters of full fine-tuning and greatly save the training cost. Finally, our retrieval performance exceeds traditional methods by 7-13\% and achieves comparable or better than full fine-tuning. Our PE-RSITR framework is both parameter-efficient and effective. 

In general, our contributions can be summarized in the following aspects.

\begin{enumerate}
 \item  We propose a novel and sophisticated PETL framework for the RS image-text retrieval task. Specifically, the proposed framework consists of the pretrained CLIP model, the MRS-Adapter, and a hybrid multi-modal contrastive learning objective.
 \item  We design a simple yet effective loss function: the hybrid multi-modal contrastive (HMMC) loss for PETL-based RS image-text retrieval. Experimental results prove that the proposed HMMC loss is effective in further improving the performance on top of the proposed MRS-Adapter.
 \item  We provide comprehensive empirical studies for the PETL-based RS image-text retrieval task. Our qualitative and quantitative results demonstrate that the proposed method is promising and of great potential for practical applications.
 \item  Extensive experiments show that our approach can achieve a parameter reduction of 98.9\% without performance sacrifice compared to full fine-tuning. Our performance exceeds traditional methods by 7-13\%. The comprehensive benchmark results are insightful for future research.
 
\end{enumerate}

This paper is organized as follows. We review the related work of RS image-text retrieval, the VLP model, and parameter-efficient transfer learning in Section \ref{sec:relatedwork}. 
In Section \ref{sec:methods}, we present our proposed PE-RSITR framework.
Evaluation methods and extensive experiment results are shown in Section \ref{sec:experiments}.
Finally, we conclude this work in Section \ref{sec:conclusion}.

\section{Related Work}
\label{sec:relatedwork}

\subsection{Remote Sensing Image-Text Retrieval}
With the development of RS vision-language cross-modal technology, RS image-text retrieval (RSITR) is becoming a major interest. RSITR can effectively verify the performance of VL modal representations. However, due to the complexity of RS image-text data, there have been limited related works.
Some works \cite{abdullah2020textrs,9154525,rahhal2020deep,cheng2021deep} employ CNN (\textit{e.g.}, VGG, ResNet) to extract RS image features and RNN (\textit{e.g.}, LSTM, BiLSTM) to extract text features.
\citet{9154525} first encoded the RS image and converted it to a caption, and finally calculates the similarity with the real captions to complete the matching. \citet{rahhal2020deep} proposed an unsupervised image-text retrieval method for RS imagery.
To reduce the occupancy and overhead of the retrieval algorithm, \citet{9594840} proposed a lightweight RS multiscale crossmodal retrieval model (LW-MCR), and designed distillation loss and semi-supervised loss to enhance the retrieval performance.
\citet{9437331} also proposed an asymmetric multimodal multi-source image retrieval method that uses the multiscale self-attention module to extract salient features of RS images and utilizes the features to guide the text representation.
In the RSITR framework based on global and local information (GaLR), \citet{9437331} indicated that RSITR should focus not only on the global features of RS images but also on the local features reflecting object relationships and saliency.
Recently, multilanguage transformer \cite{9925582} demonstrated that loading CLIP pre-training model \cite{pmlr-v139-radford21a} can achieve promising performance on RS image-text retrieval.

\subsection{Vision-Language Pre-training Model}
Large-scale VLP models are developing rapidly and have shown encouraging results on various downstream tasks \cite{chen2023vlp}. 
According to the encoder type, VLP models are mainly classified into fusion encoder and dual encoder \cite{du2022survey}. 
The fusion encoder takes image and text features as input and uses some fusion methods for VL interaction. The fusion encoder is mainly classified into single-stream and dual-stream structures. The single-stream structure (\textit{e.g.}, OSCAR \cite{li2020oscar}, XGPT \cite{xia2021xgpt}, SimVLM \cite{wang2022simvlm}) concatenates multimodal features and uses the transformer encoder in a unified framework. However, the single-stream performs the self-attention directly on two modalities, ignoring the inter-modal interaction. Therefore the dual-stream structure (\textit{e.g.}, ViLBERT \cite{NEURIPS2019_c74d97b0}, ALBEF \cite{li2021align}) performs the cross-attention using the transformer decoder.
The fusion encoder relies on a large transformer for VL interaction modeling but the inference process can be very slow in solving matching tasks such as image-text retrieval.
In contrast, the dual encoder (\textit{e.g.}, CLIP \cite{pmlr-v139-radford21a}, ALIGN \cite{pmlr-v139-jia21b}) uses some simple methods for VL interaction modeling and calculates similarity scores after projecting the image and text features into the same semantic subspace.
This method is more efficient for retrieval tasks, \textit{e.g.} CLIP shows amazing results for the image-text retrieval, but does not work well when dealing with VL understanding tasks.
The current pre-training models for RS image research have also been greatly developed \cite{9782149}. \citet{zhang2022consecutive} proposes a transfer learning method from natural scenes to the RS domain, which can achieve good results in many tasks (\textit{e.g.}, scene classification, object detection, and land cover classification). Unfortunately, there is no sufficiently large and uniform RS image-text dataset to support VLP models for the RS domain. 

\subsection{Parameter-Efficient Transfer Learning}
Existing methods of PETL are broadly divided into two families, \textit{i.e.} prompt learning and adapter, and are summarized in detail as follows.
\subsubsection{Prompt Learning}
Prompt learning \cite{li2021prefix, lester2021power} was first proposed in natural language processing (NLP). When fine-tuning large language models, task-specific learnable vectors are added to the input. Unlike full fine-tuning, prompt learning can significantly reduce storage and computational costs and can achieve comparable performance. 
Prompt learning has been applied to computer vision (CV). Visual Prompt Tuning \cite{10.1007} (VPT) can efficiently fine-tune large-scale transformer models in vision.
For CLIP-based image classification, \citet{Zhou2022CoOp} added continuous learnable prompts on text labels for context optimization (CoOp) and then proposed to generate prompts using image features for conditional context optimization \cite{9879913} (CoCoOp). 
\citet{zang2022unified} further combined the advantages of text prompt learning and visual prompt learning to propose a tiny network to jointly optimize prompt learning for different modalities. MaPLe \cite{khattak2022maple} used the V-L coupling function to generate visual prompts on textual prompts to adapt both language and vision branches simultaneously.
However, there are few prompt learning methods for multimodal tasks. CPT \cite{yao2021cpt} proposed a new prompt learning paradigm for visual grounding by adding color prompts to text and images, respectively. CPT used the color-masked token of the target region and color text prompts to ground the object.

\subsubsection{Adapter}
Since \citet{houlsby19a} proposed the adapter module to fine-tune large pretrained models in NLP, many improved methods \cite{adapterfusion,NEURIPS2021_081be9fd,Hyperparameter,adapterdrop} have shown good performance on NLP tasks. 
Adapter is a lightweight plug-and-play module. The pre-trained model is frozen during fine-tuning and the parameters of the adapter are updated.
Recently, the adapter is widely used to fine-tune pre-trained models in CV.
\citet{chen2022conv} proposed Conv-Adapter which replaces the original linear layers with convolutional layers, making it possible to efficiently fine-tune largescale ConvNets. Convpass \cite{jie2022convolutional} is also composed of convolutional layers, but it is the adapter for vision transformer. 
AdapterFormer \cite{chen2022adaptformer} is the adapter based on the original linear layers for vision transformer, and it can adapt both image and video tasks efficiently. 
\citet{pan2022stadapter} proposed a new Spatio-Temporal Adapter (ST-Adapter) to accomplish PETL from image models to video tasks.
With the appearance of CLIP, the adapter combined with CLIP further shows superior performance. CLIP-Adapter \cite{10.48550} employs an additional bottleneck layer with residual connections at the end of the image and text branches, respectively. \citet{zhang2022tip} proposed a training-free Tip-Adapter for CLIP for few-shot classification. 
SVL-Adapter \cite{pantazis2022svl} combines the complementary advantages of CLIP and self-supervised representation learning for image classification that are significantly different from common images.
Previous works had focused on unimodal tasks and less on multimodal tasks. Although VL-Adapter \cite{Sung_2022_CVPR} is proposed for image-text and video-text multimodal tasks, it utilizes CLIP for image encoding and uses Adapter only in the language model.
The latest Cross-Modal Adapter \cite{Jiang10.48550} and UniAdapter \cite{2302.06605} both propose cross-modal interaction mechanism and have shown good performance in multimodal tasks such as image/video text retrieval and visual question answering.
 
\section{Methodology}
\label{sec:methods}
\begin{figure*}[t]
	\centering
	\includegraphics[width=0.95\linewidth]{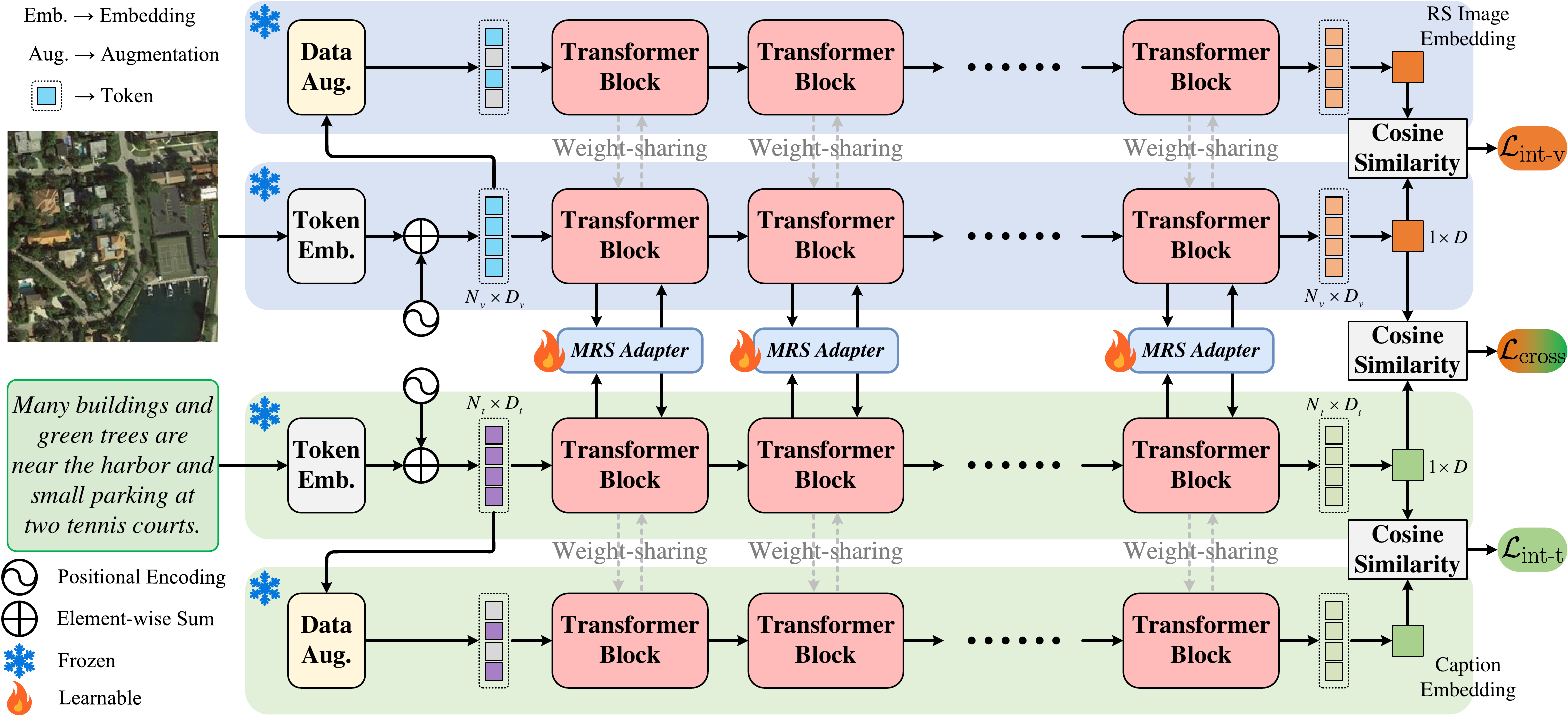}
	\caption{Overall architecture of our proposed novel and sophisticated PE-RSITR framework. It mainly consists of three parts: the frozen CLIP pre-training backbone, the multimodal remote sensing adapter (MRS-Adapter) with cross-modal interaction, and a hybrid multi-modal contrastive (HMMC) loss.}
	\label{fig:method}
\end{figure*}

To capitalize a large VLP model of natural scenes for RS vision-language tasks with domain differences, such as RSITR, the intrinsic gap in different domains need to be filled. This section illustrates our proposed PE-RSITR framework in detail, and the overall framework is shown in Fig. \ref{fig:method}.

\subsection{Preliminary}
\label{sec:Preliminary}
In this subsection, we briefly describe how to process RS images and caption embeddings through the CLIP pre-training model and introduce the basic structure of the adapter.

\textbf{Multimodal Encoder.}
We use the CLIP pre-training model as the primary multimodal encoder, including two branches of image encoder (ViT-B/32) and text encoder with the same structure.
Concretely, given a cross-modal RS image-query dataset $\mathcal{O} =\left \{ \left ( i_{n} , t_{n}  \right )  \right \} _{n=1}^{N} $, where there are \textit{N} pairs of image-text positive pairs $(i_{n} , t_{n})$. To simplify the notations, we denote $\boldsymbol{I}\in \mathbb{R}^{H\times W\times 3}$ and $T=\left \{ w_{m} \right \}_{m=1}^{M}$ (\textit{M} is the sentence length) as single instances of RS image and query text modality, respectively, where ${H\times W\times 3}$ denotes the size of the RS image and $w_{m}$ represents the $m$-th word. 
The basic architecture of the image encoder is shown in the blue background branch in Fig. \ref{fig:method}.
First, a convolution layer is used to generate the patch tokens $\boldsymbol{I}_{patch} \in \mathbb{R}^{\frac{H}{s} \cdot  \frac{W}{s}\times D_{v}}$, where $s$ is the stride of the backbone network and also the patch size. An additional classification (CLS) token $\boldsymbol{I}_{cls} \in \mathbb{R}^{1 \times D_{v}}$ is added to the token sequence, and the vision transformer adds the positional embedding $\boldsymbol{I}_{pos} \in \mathbb{R}^{{N_{v}}\times D_{v}}$ into each token. The RS image tokens are defined as
\begin{equation}\label{eq:I_token}
    \boldsymbol{I}_{0} =\left [ \boldsymbol{I}_{cls}\text{;} \  \boldsymbol{I}_{patch}   \right ] + \boldsymbol{I}_{pos},
\end{equation}
where $\boldsymbol{I}_{0} \in \mathbb{R}^{{N_{v}}\times D_{v}}$, $N_{v}=1+\frac{H}{s} \cdot \frac{W}{s}$, and $D_{v}$ is the hidden dimension of the vision transformer. Afterward, the RS image tokens $\boldsymbol{I}_{0}$ pass through 12 layers of stacked transformer blocks. The $l$-th transformer block can be represented as
\begin{align}
   \hat{\boldsymbol{I}}_{l} &= \text{MHA}  ( \text{LN} ( \boldsymbol{I}_{l-1}  )  )  + \boldsymbol{I}_{l-1},\\
   \boldsymbol{I}_{l} &= \text{MLP}  ( \text{LN} ( \hat{\boldsymbol{I}}_{l} )  ) + \hat{\boldsymbol{I}}_{l},
\end{align}
where $\hat{\boldsymbol{I}}_{l}$ and $\boldsymbol{I}_{l}$ respectively indicate the output of multi-head attention (MHA) and feed-forward network (FNN) modules.
Finally, the CLS token is used as the global RS image embedding $\boldsymbol{v}$ which is linearly projected into the $D$-dimensional cross-modal semantic space and then $L_{2}$ normalized.

The green background branch in Fig. \ref{fig:method} is the text encoder. The text encoder performs token embedding similarly. First, the caption is tokenized using lower-cased byte pair encoding (BPE), denoted as $\boldsymbol{T}_{token} \in \mathbb{R}^{{N_{t}}\times D_{t}}$. The token sequence for each caption starts with a $[BOS]$ token and ends with a $[EOS]$ token. Afterwards, the positional embedding $\boldsymbol{T}_{pos} \in \mathbb{R}^{{N_{t}}\times D_{t}}$ is added to each token. The text tokens are defined as
\begin{equation}\label{eq:T_token}
    \boldsymbol{T}_{0} = \boldsymbol{T}_{token} + \boldsymbol{T}_{pos},
\end{equation}
where $\boldsymbol{T}_{0} \in \mathbb{R}^{{N_{t}}\times D_{t}}$ and $D_{t}$ is the hidden dimension of the text transformer. After the word embedding, the tokens $\boldsymbol{T}_{0}$ are sent to 12 layers of stacked transformer blocks. The $l$-th transformer block can be represented as
\begin{align}
   \hat{\boldsymbol{T}}_{l} &=\text{MHA} ( \text{LN} ( \boldsymbol{T}_{l-1}  )   )  + \boldsymbol{T}_{l-1},\\
   \boldsymbol{T}_{l} &= \text{MLP} ( \text{LN} ( \hat{\boldsymbol{T}}_{l}  )   ) + \hat{\boldsymbol{T}}_{l},
\end{align}
where $\hat{\boldsymbol{T}}_{l}$ and $\boldsymbol{T}_{l}$ respectively indicate the output of MHA and FNN modules.
Finally, the highest layer of the transformer at the $[EOS]$ token is used as the global caption embedding $\boldsymbol{t}$ which is linearly projected into the $D$-dimensional cross-modal semantic space and then $L_{2}$ normalized. The only difference between the image encoder and the text encoder is that the hidden dimension $D_{v}$=768 while $D_{t}$=512.

\textbf{Adapter.}
Inspired by the success of Adapter \cite{houlsby19a} in NLP, more and more adapter-based methods have shown promising performance in both CV and VL tasks. The adapter consists of a bottleneck that contains few parameters relative to the original pre-training model. Specifically, the adapter first uses a down-projection linear layer with parameters $\boldsymbol{W}_{down} \in \mathbb{R}^{d \times \hat{d} }$ ($\hat{d} \ll d$) to project the input features onto a low-dimensional representation. Then a non-linear activation function is used, commonly the ReLU activation function. Finally, an up-projection linear layer with parameters $\boldsymbol{W}_{up} \in \mathbb{R}^{\hat{d} \times d }$ projects the features back to the input size before adding the skip connection. Formally, given an input feature $\boldsymbol{X} \in \mathbb{R}^{N_{in} \times d}$, the adapted feature $\tilde{\boldsymbol{X}} \in \mathbb{R}^{N_{in} \times d}$ can be calculated as
\begin{align}
   \boldsymbol{X}^{down} &= \text{ReLU}(\boldsymbol{X} \cdot \boldsymbol{W}_{down}),\\
   \text{Adapter}(\boldsymbol{X}) &= \tilde{\boldsymbol{X}} = s\cdot \boldsymbol{X}^{down} \cdot \boldsymbol{W}_{up} + \boldsymbol{X},
\end{align}
where $s$ is a scalar scale factor that controls the effect of the adapter.
The original adapter scheme is inserted sequentially into the MHA and FFN of the transformer block. The feature adaptation process at the $l$-th layer can be written as
\begin{align}
   \hat{\boldsymbol{X}}_{l} &= \text{Adapter}( \text{MHA}\left ( \text{LN}\left ( \boldsymbol{X} \right )  \right ))  + \boldsymbol{X},\\
    \boldsymbol{X}_{l} &= \text{Adapter}(\text{MLP} ( \text{LN} ( \hat{\boldsymbol{X}}_{l} )  ))  + \hat{\boldsymbol{X}}_{l}.
\end{align}

\subsection{MRS-Adapter}
\label{sec:MRS-Adapter}
\begin{figure}[t]	
	\centering
	\includegraphics[width=0.95\linewidth]{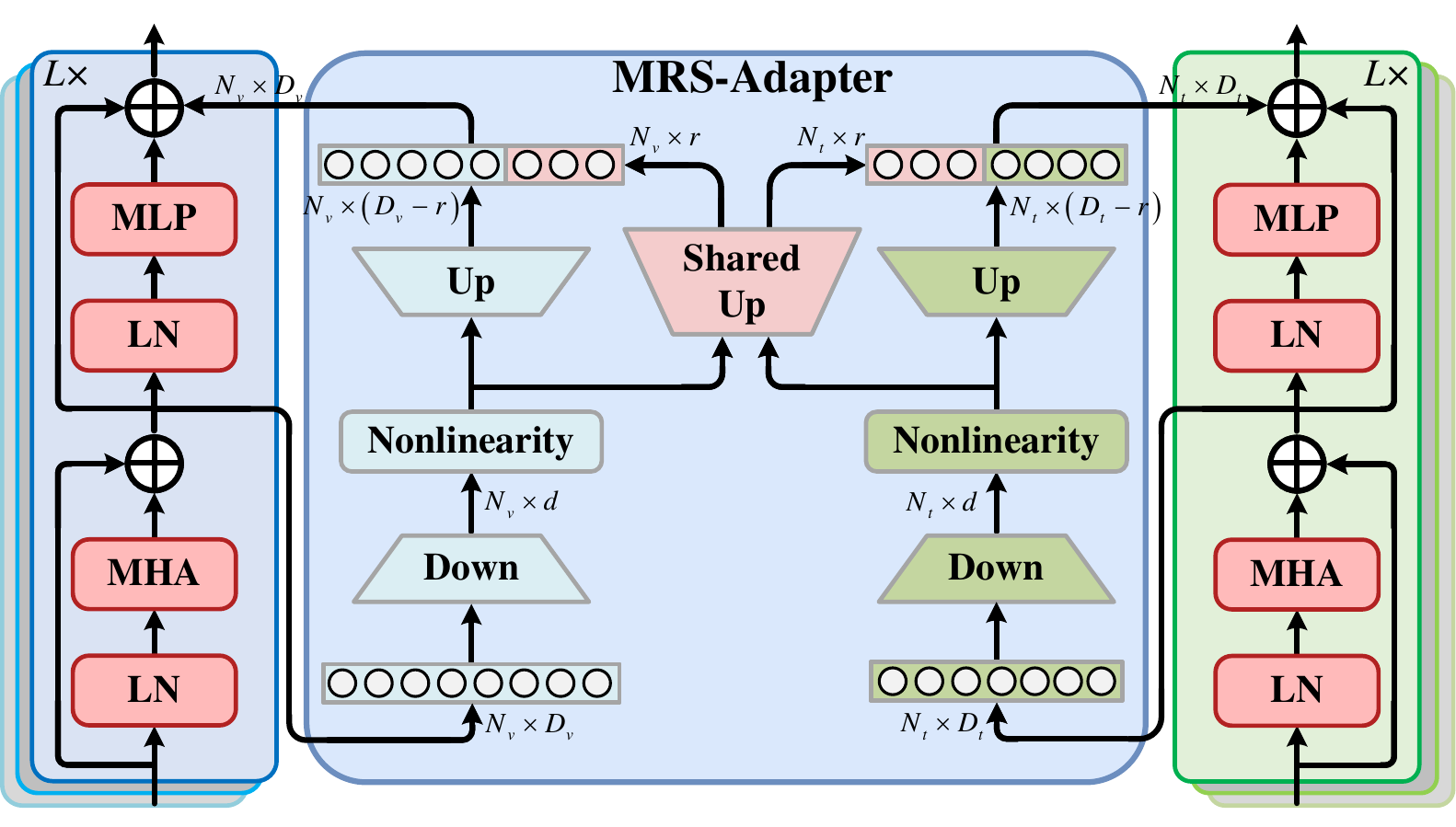}
	\caption{The implementation details of our MRS-Adapter. MRS-Adapter is inserted in parallel with the FFN module and adds a modality-shared up-projection to connect the original modality-specific up-projection linear layer for cross-modal weight sharing.}
	\label{fig:MRS-Adapter}
\end{figure}

Various CLIP-based adapter methods show great potential for VL tasks. The core of extending CLIP to the RS domain to accomplish RSITR lies in exploring the VL expert knowledge in the RS domain efficiently while appropriately inheriting the VL prior knowledge structure of the natural scene domain. This work is still under-explored. If the adapter is directly extended from NLP to both modalities of VL, it can only lead to sub-optimal results due to the lack of cross-modal interactions. This point was also verified in the recent works \cite{Jiang10.48550,2302.06605}. Therefore, we attempt to design an adapter that can share knowledge between RS image modality and text modality without increasing parameters. 
Finally, we found that the same cross-modal interaction as Cross-Modal Adapter is the most effective way and can reduce parameters.
Our MRS-Adapter is extremely similar to Cross-Modal Adapter, but our scheme is more concise. MRS-Adapter has no skip connection and is only connected in parallel with the FFN module, which can further reduce the number of adapters. The specific structure is shown in Fig. \ref{fig:MRS-Adapter}. 
Formally, the input of $l$-th layer is the $\hat{\boldsymbol{I}}_{l} \in \mathbb{R}^{N_{v} \times D_{v}}$ and $\hat{\boldsymbol{T}}_{l} \in \mathbb{R}^{N_{t} \times D_{t}}$ of the MHA module output, and the adapted features can be obtained as follows:
\begin{align}
   \hat{\boldsymbol{I}}_{l}^{down} &= \text{ReLU}(\hat{\boldsymbol{I}}_{l} \cdot \boldsymbol{W}_{down}^{v}),\\
   \hat{\boldsymbol{I}}_{l}^{\text{MRS-Adapter}} &= [\hat{\boldsymbol{I}}_{l}^{down} \cdot \boldsymbol{W}_{up}^{v}\text{;} \ \hat{\boldsymbol{I}}_{l}^{down} \cdot \boldsymbol{W}_{up}^{share}],\\
   \hat{\boldsymbol{T}}_{l}^{down} &= \text{ReLU}(\hat{\boldsymbol{T}}_{l} \cdot \boldsymbol{W}_{down}^{t}),\\
    \hat{\boldsymbol{T}}_{l}^{\text{MRS-Adapter}} &= [\hat{\boldsymbol{T}}_{l}^{down} \cdot \boldsymbol{W}_{up}^{t}\text{;} \ \hat{\boldsymbol{T}}_{l}^{down} \cdot \boldsymbol{W}_{up}^{share}],
\end{align}
where $d \ll D_{v}$, $d \ll D_{t}$, $0< r < D_{v}$, $0< r < D_{t}$, $\boldsymbol{W}_{down}^{v} \in \mathbb{R}^{D_{v}  \times d}$ and $\boldsymbol{W}_{down}^{t} \in \mathbb{R}^{D_{t}  \times d}$ are the modality-specific down-projection weights of two branches, $\boldsymbol{W}_{up}^{share} \in \mathbb{R}^{d  \times r}$ is modality-shared weights, and $\boldsymbol{W}_{up}^{v} \in \mathbb{R}^{d \times D_{v}-r }$ and $\boldsymbol{W}_{up}^{t} \in \mathbb{R}^{d \times D_{t}-r }$ are the modality-specific up-projection weights of two branches. 
Finally, the MRS-Adapter is connected in parallel with the FFN module at each layer of both branches, and the feature adaptation process at $l$-th layer can be written as
\begin{align}
   \hat{\boldsymbol{I}}_{l} &=\text{MHA}\left ( \text{LN}\left ( \boldsymbol{I}_{l-1} \right )  \right )  + \boldsymbol{I}_{l-1},\\
   \boldsymbol{I}_{l} &= \text{MLP} ( \text{LN} ( \hat{\boldsymbol{I}}_{l}  )   ) + \hat{\boldsymbol{I}}_{l} + \hat{\boldsymbol{I}}_{l}^{\text{MRS-Adapter}},\\
   \hat{\boldsymbol{T}}_{l} &=\text{MHA}\left ( \text{LN}\left ( \boldsymbol{T}_{l-1} \right )  \right )  + \boldsymbol{T}_{l-1},\\
    \boldsymbol{T}_{l} &= \text{MLP} ( \text{LN} ( \hat{\boldsymbol{T}}_{l}  )   ) + \hat{\boldsymbol{T}}_{l} +\hat{\boldsymbol{T}}_{l}^{\text{MRS-Adapter}}.
\end{align}

MRS-Adapter adding an $r$-dimensional linear layer for weight sharing can directly reduce $d \times r$ parameters. The shared up-projection enables the fine-grained information of RS image modality and text modality to interact, which can enhance the RS vision language modality representation. MRS-Adapter can learn the VL knowledge specific to the RS domain and efficiently extend the natural scene domain to the RS domain.

\subsection{HMMC Learning Objective}
\label{sec:loss_fun}
In image-text retrieval tasks, the bi-directional triplet loss established by \citet{VSE} has become the mainstream loss function. The bi-directional triplet loss can pull the distance between this sample and the positive sample of another modality closer while pushing the distance between it and the negative sample of another modality farther.
However, this framework can only constrain the inter-modal samples and does not consider the intra-modal samples. In particular, RS images are characterized by extremely high intra-class similarity, and the visualization results of the textual similarity in the literature \cite{9437331} show that the textual similarity is also extremely high. Therefore, RSITR often results in the error of retrieving misalignment of similar RS images or captions. To cope with the problem, token-level data augmentation is adopted to construct intra-modal positive pairs for RS images and texts. We design a simple yet effective hybrid multi-modal contrastive loss function. The framework of inter-modal and intra-modal cooperative constraints is shown in Fig. \ref{fig:method}. Inspired by contrastive learning \cite{gao2021simcse}, the method of data augmentation is to adopt the simple random dropout.

\textbf{Intra-Modal Constraints.}
The bi-directional triplet loss with hard negatives is used, and the similarity between sample pairs is measured by cosine similarity.
For RS image modality, data augmentation is performed after token embedding, and a token-level positive pair $(\boldsymbol{I}_{0}, \boldsymbol{I}_{0}^{+})$ is obtained for each RS image $\boldsymbol{I}$. The process is denoted as
\begin{equation}\label{eq:drop_v}
    \boldsymbol{I}_{0}^{+} =\text{random\_dropout}\left ( \boldsymbol{I}_{0}  \right ).
\end{equation}

Then we perform the intra-modal constraint of the RS image to pull the distance from the query image and other similar images. 
We denote the final embedding of $\boldsymbol{I}_{0}^{+}$ as $\boldsymbol{v}^{+}$.
For a positive pair $(\boldsymbol{v},\boldsymbol{v}^{+})$, the intra-modal triplet loss we adopt is:

\begin{equation}\label{eq:loss-v}
    \begin{split}
    \mathcal{L}_{\text{intra-v}}(\boldsymbol{v}, \boldsymbol{v}^{+})
    & = \sum_{\widehat{\boldsymbol{v}}^{+}}[\alpha_{v}- \text{cos}(\boldsymbol{v}, \boldsymbol{v}^{+}) + \text{cos}(\boldsymbol{v}, \widehat{\boldsymbol{v}}^{+})]_{+} \\ 
    & + \sum_{\widehat{\boldsymbol{v}}}[\alpha_{v}- \text{cos}(\boldsymbol{v}, \boldsymbol{v}^{+}) + \text{cos}(\widehat{\boldsymbol{v}}, \boldsymbol{v}^{+})]_{+}, 
    \end{split}
\end{equation}
where $\alpha_{v}$ is the margin of the intra-modal constraint of the RS image and $[x]_{+}=max(x,0)$. Similarly, the text branch also uses random dropout to generate a token-level positive pair $(\boldsymbol{T}_{0}, \boldsymbol{T}_{0}^{+})$ for each caption $\boldsymbol{T}$, calculated as follows:
\begin{equation}\label{eq:drop_t}
    \boldsymbol{T}_{0}^{+} =\text{random\_dropout}\left ( \boldsymbol{T}_{0}  \right ).
\end{equation}

Then we perform the intra-modal constraint of the caption text to pull the distance from the query caption and other similar captions. We denote the final embedding of $\boldsymbol{T}_{0}^{+}$ as $\boldsymbol{t}^{+}$.
For a positive pair $(\boldsymbol{t},\boldsymbol{t}^{+})$, the intra-modal triplet loss is:
\begin{equation}\label{eq:loss-t}
    \begin{split}
    \mathcal{L}_{\text{intra-t}}(\boldsymbol{t}, {\boldsymbol{t}}^{+})
    & = \sum_{\widehat{\boldsymbol{t}}^{+}}[\alpha_{t}- \text{cos}(\boldsymbol{t}, \boldsymbol{t}^{+}) + \text{cos}(\boldsymbol{t}, \widehat{\boldsymbol{t}}^{+})]_{+} \\ 
    & + \sum_{\widehat{\boldsymbol{t}}}[\alpha_{t}- \text{cos}(\boldsymbol{t}, \boldsymbol{t}^{+}) + \text{cos}(\widehat{\boldsymbol{t}}, \boldsymbol{t}^{+})]_{+}, 
    \end{split}
\end{equation}
where $\alpha_{t}$ is the margin of the intra-modal constraint of the caption text.

\textbf{Cross-Modal Constraint.}
The multi-modal alignment of RS image-text is promoted by relying on the global similarity of RS image-text. We compute the cross-modal constraint loss with
\begin{equation}\label{eq:loss}
    \begin{split}
    \mathcal{L}_{\text{cross}}(\boldsymbol{v}, \boldsymbol{t})
    & = \sum_{\widehat{\boldsymbol{t}}}[\lambda- \text{cos}(\boldsymbol{v}, \boldsymbol{t}) + \text{cos}(\boldsymbol{v}, \widehat{\boldsymbol{t}})]_{+} \\ 
    & + \sum_{\widehat{\boldsymbol{v}}}[\lambda- \text{cos}(\boldsymbol{v}, \boldsymbol{t}) + \text{cos}(\widehat{\boldsymbol{v}}, \boldsymbol{t})]_{+},
    \end{split}
\end{equation}
where $\lambda$ is the margin of the cross-modal constraint.

\textbf{Overall Objective.}
By combining hard-negative-based intra-modal constraints loss with cross-modal constraints loss, we obtain the HMMC loss:
\begin{equation}
    \mathcal{L}(\boldsymbol{v}, \boldsymbol{t}) = \mathcal{L}_{\text{cross}}(\boldsymbol{v}, \boldsymbol{t}) + \mathcal{L}_{\text{intra-v}}(\boldsymbol{v}, {\boldsymbol{v}}^{+})+ \mathcal{L}_{\text{intra-t}}(\boldsymbol{t}. {\boldsymbol{t}}^{+}).
\end{equation}

\section{Experiments}
\label{sec:experiments}
In this section, we first describe the dataset, evaluation metrics, and experimental setup details in Section \ref{sec:evaluation} and Section \ref{sec:implementation}. 
Further, Section \ref{sec:comparisons} introduces the SOTA approaches and provides comparisons of retrieval performance. Section \ref{sec:analysis} conducts result analyses.
In Section \ref{sec:ablation}, we perform sufficient ablation experiments.
Finally, we present some visualization results to further analyze in Section \ref{sec:qualitative}.

\subsection{Dataset and Evaluation Metrics}
\label{sec:evaluation}
We evaluate our proposed PE-RSITR framework on the three widely used RS image-text datasets: RSICD \cite{8240966}, RSITMD \cite{9437331}, and UCM \cite{7546397}. RSICD is the dataset with the largest number of samples, while RSITMD is the dataset with more fine-grained captions and more challenges. UCM requires the model to be robust because of the small numbers.

Two evaluation metrics Recall at $K$ ($R@K$, $K$=1, 5, and 10) and mean recall (mR) are exploited to assess our model.
$R@K$ aims to calculate the ratio of queries that successfully retrieve the ground truth as one of the first $K$ results. $mR$ represents the average of $R@K$ for both the text retrieval and image retrieval, which evaluates the overall retrieval performance and can be formulated in the equation below,
\begin{equation}
     mR=(\underset{\text{Text \ retrieval}}{\underbrace{R@1+R@5+R@10}}  + \underset{\text{Image \ retrieval}}{\underbrace{R@1+R@5+R@10} })/6.
 \end{equation}


\subsection{Implementation Details}
\label{sec:implementation}
All experiments in this work are conducted on one NVIDIA RTX 3090 24GB GPU.
We follow the data partitioning approach of \citet{9437331} and use 80\%, 10\%, and 10\% of the dataset as the training set, validation set, and test set, respectively.
For the RS image, we resize the image size to a fixed size of 224$\times$224 for training. 
We set the dimension $d$ and $r$ to 64 and the probability of random dropout to 0.2. The margin $\lambda$, $\alpha_{v}$, $\alpha_{t}$ are set to 0.2 for the triplet loss calculation.
We set the initial learning rate of our network to 0.0002 for trained parameters and weight decay by 0.7 every 20 epochs. 
During training, we adopt the Adam optimizer to train our network with a batch size of 16 for 30 epochs.
To make the experiment more convincing, we follow the works in GaLR \cite{9745546} and MCRN \cite{YUAN2022103071} to conduct the experiments and report results. We leverage k-fold cross-validation to obtain an average result, and k is set to 5.

\subsection{Comparisons with State-of-the-art Methods}
\label{sec:comparisons}
In this experiment, we comprehensively compare our proposed method with traditional cross-modal retrieval methods and CLIP-based methods. 

\textbf{Traditional methods:}
Following the previous literature, we also compare the proposed method with the progressive image-text retrieval models (VSE++ \cite{VSE}, SCAN \cite{Lee_2018_ECCV}, CAMP \cite{Wang_2019_ICCV}, MTFN \cite{10.1145/3343031.3350875}, LW-MCR \cite{9594840}, AMFMN \cite{9437331}, GaLR \cite{9745546}) on three RS image-text datasets. For these methods, we use the results in three literature \cite{9594840,9437331, 9745546}. In addition, we have added two latest RS image-text retrieval methods.

\begin{itemize}
    \item \textit{MCRN} \cite{YUAN2022103071}:
    MCRN constructs a multi-source cross-modal retrieval network capable of image modality, text modality, and audio modality alignment based on shared networks of pattern memory and generative adversarial theory.
    \item \textit{CABIR} \cite{app122312221}:
    CABIR proposes a cross-attention model based on region-level semantic features of RS images, with textual semantics to allocate weights and filter redundant features for image regions.
\end{itemize}

\textbf{CLIP-based methods:}
Single-Language \cite{9925582} is loaded with CLIP pre-training parameters for training. In addition, we benchmark extensive efficient and commonly used PETL approaches in our PE-RSITR task. In order to assess the merits of our proposed method, we report our performance and compare it with the following methods. 

\begin{figure}[t]	
	\centering
	\includegraphics[width=1\linewidth]{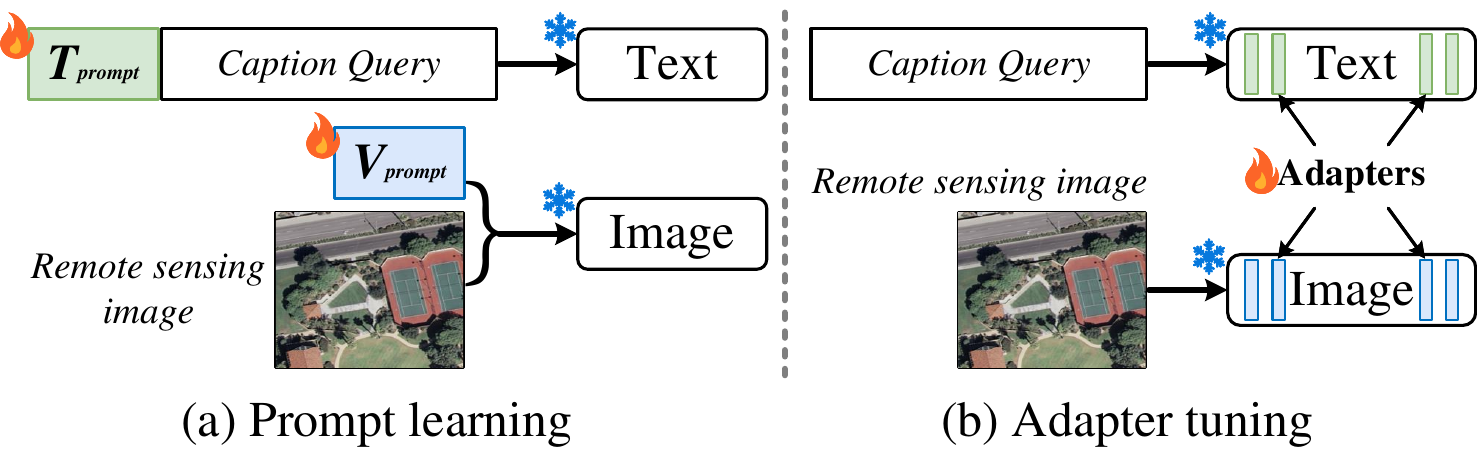}
	\caption{(a): Prepending a sequence of learnable prompt tokens to the input tokens of the visual or text encoder. Only these added prompt tokens are updated during fine-tuning; (b): Insert a lightweight adapter into the visual or text encoder and update the parameters of the adapter during fine-tuning.}
	\label{fig:Comparison}
\end{figure}

\begin{itemize}
    \item \textit{Zero-shot CLIP} \cite{pmlr-v139-radford21a}: Testing directly on the sample of the unseen RS domain.
    \item \textit{Linear Probe}: Adding an extra linear layer on top of each of the two branches of the backbone and freezing all the parameters except the parameters in the linear layer.
    \item \textit{Full Fine-tuning}: Fully updating all the parameters. 
    In the experiments of the RSICD dataset, Full Fine-tuning utilizes the weight of the CLIP-rsicd model\textsuperscript{\ref{note1}}.
    \item \textit{Prompt Learning}: As shown in Fig. \ref{fig:Comparison}(a), prepending a sequence of learnable prompt tokens to the input tokens, and only these added prompt tokens are updated during fine-tuning. 
     Specifically, we compare CoOp \cite{Zhou2022CoOp} (added in the text tokens) and VPT \cite{10.1007} (added in the visual tokens). Following \citet{Jiang10.48550}, we applied both the visual and text tokens, called VL-Prompt.
    \item \textit{Adapter Tuning}:
    Adapter is a lightweight plug-and-play module. The pre-trained model is frozen during fine-tuning and the parameters of adapters are updated, as shown in Fig. \ref{fig:Comparison}(b).
    CLIP-Adapter \cite{10.48550} employs an additional bottleneck layer to learn new features and make residual connections.
    Adapter \cite{houlsby19a} is for the language transformer, Convpass \cite{jie2022convolutional} is the convolutional adapter for ViT, and AdaptFormer \cite{chen2022adaptformer} is for adapting ViT to different image and video tasks.
    UniAdapter \cite{2302.06605} and Cross-Modal Adapter \cite{Jiang10.48550} are multimodal adapters. UniAdapter cannot be used directly due to the different dimensions of the visual branch and text branch. We use a linear layer to project visual features to 512 dimensions.
\end{itemize}

\begin{table*}[]
\centering
\caption{
The retrieval performance of state-of-the-art methods on RSICD test set. The best performance of traditional methods is with bold.
\textcolor{red}{Red} and \textcolor{blue}{blue} represent the best and second performance in CLIP-based and transfer learning methods.
}
\begin{tabular}{lcccccccccc}
\hline
     &   &   &    & \multicolumn{3}{c}{\textbf{Text retrieval}}                                                                       & \multicolumn{3}{c}{\textbf{Image retrieval}}                                                                      &                               \\
\multirow{-2}{*}{\textbf{Methods}}  &  \multirow{-2}{*}{\textbf{Reference}}    &  \multirow{-2}{*}{\textbf{Backbone (image/text)}} &  \multirow{-2}{*}{\textbf{Params}}   & \textbf{R@1}                  & \textbf{R@5}                  & \textbf{R@10}                 & \textbf{R@1}                  & \textbf{R@5}                  & \textbf{R@10}                 & \multirow{-2}{*}{\textbf{mR}} \\ \hline
\multicolumn{11}{c}{\textbf{\textit{\textbf{Traditional methods}}}} \\ \hline
VSE++ \cite{VSE}      & \textit{BMVC'18}    & ResNet18, GRU  &  15.78M             & 3.38                          & 9.51                          & 17.46                         & 2.82                          & 11.32                         & 18.10                         & 10.43                         \\
SCAN t-i \cite{Lee_2018_ECCV}   & \textit{ECCV'18}     & Faster R-CNN, biGRU  &  13.68M            & 4.39                          & 10.90                         & 17.64                         & 3.91                          & 16.20                         & 26.49                         & 13.25                         \\
SCAN i-t \cite{Lee_2018_ECCV}   &   \textit{ECCV'18}    & Faster R-CNN, biGRU  &  13.68M             & 5.85                          & 12.89                         & 19.84                         & 3.71                          & 16.40                         & 26.73                         & 14.23                         \\
CAMP-triplet \cite{Wang_2019_ICCV}  &  \textit{ICCV'19}   &Faster R-CNN, biGRU   & 27.03M          & 5.12                          & 12.89                         & 21.12                         & 4.15                          & 15.23                         & 27.81                         & 14.39                         \\
CAMP-bce \cite{Wang_2019_ICCV}     &  \textit{ICCV'19}     & Faster R-CNN, biGRU  & 27.03M          & 4.20                          & 10.24                         & 15.45                         & 2.72                          & 12.76                         & 22.89                         & 11.38                         \\
MTFN \cite{10.1145/3343031.3350875}   &   \textit{ACM MM'19}    &Faster R-CNN, biGRU   & 77.90M                 & 5.02                          & 12.52                         & 19.74                         & 4.90                          & 17.17                         & 29.49                         & 14.81                         \\
LW-MCR-b \cite{9594840}      &  \textit{TGRS'22}  & SqueezeNet, -  &  1.65M            & 4.57                          & 13.71                         & 20.11                         & 4.02                          & 16.47                         & 28.23                         & 14.52                         \\
LW-MCR-d \cite{9594840}      &  \textit{TGRS'22}  & SqueezeNet, -   & 1.65M           & 3.29                          & 12.52                         & 19.93                         & 4.66                          & 17.51                         & 30.02                         & 14.66                         \\
LW-MCR-u \cite{9594840}    &  \textit{TGRS'22}  & SqueezeNet, -   &  1.65M             & 4.39                          & 13.35                         & 20.29                         & 4.30                          & 18.85                         & 32.34                & 15.59                         \\
AMFMN-soft \cite{9437331}  &  \textit{TGRS'22}   & ResNet18, biGRU  & 35.94M              & 5.05                          & 14.53                         & 21.57                         & 5.05                          & 19.74                         & 31.04                        & 16.02                         \\
AMFMN-fusion \cite{9437331}   &   \textit{TGRS'22}   & ResNet18, biGRU  & 35.94M          & 5.39                          & 15.08                         & 23.40                        & 4.90                          & 18.28                         & 31.44                         & 16.42                         \\
AMFMN-sim \cite{9437331}  &   \textit{TGRS'22} & ResNet18, biGRU   & 35.94M              & 5.21                          & 14.72                         & 21.57                         & 4.08                          & 17.00                         & 30.60                         & 15.53                         \\
 MCRN \cite{YUAN2022103071}   &  \textit{JAG'22}    & ResNet18, biGRU  & 52.35M          & 6.59                         & 19.40                         & 30.28                & 5.03                         & 19.38                         & 32.99                         & 18.95                         \\
 CABIR \cite{app122312221}   &  \textit{AS'22}  & ResNet152, BERT+biGRU  & -          & \textbf{8.59}                         & 16.27                         & 24.13                & \textbf{5.42}                         & \textbf{20.77}                        & \textbf{33.58}                         & 18.12                         \\
GaLR w/o MR \cite{9745546}      &  \textit{TGRS'22} & ResNet18, biGRU  & 46.89M          & 6.50                          & 18.91                         & 29.70                         & 5.11                 & 19.57                & 31.92                         & 18.62                         \\
GaLR with MR \cite{9745546}     & \textit{TGRS'22}  & ResNet18, biGRU  & 46.89M         & 6.59                 & \textbf{19.85}                & \textbf{31.04}                & 4.69                          & 19.48                         & 32.13                         & \textbf{18.96}                \\ \hline
\multicolumn{11}{c}{\textbf{\textit{\textbf{CLIP-based methods}}}} \\ \hline
Single Language \cite{9925582}  & \textit{JSTARS'22}  & CLIP(ViT-B-32)  & 151M         & 10.70                         & 29.64                         & 41.53                         & 9.14                          & 28.96                         & 44.59                         & 27.42                         \\
Zero-shot CLIP \cite{pmlr-v139-radford21a}    &  \textit{ICML'21}  &  CLIP(ViT-B-32)  & 0.00M        & 6.77                          & 15.37                         & 23.15                         & 5.01                          & 15.75                         & 24.21                        & 15.04                         \\
Linear probe \cite{pmlr-v139-radford21a}    &  \textit{ICML'21}    &  CLIP(ViT-B-32)  & 1.05M         & 8.46                          & 24.41                         & 37.72                         & 7.81                          & 25.89                         & 42.47                         & 24.46                         \\
Full fine-tuning\textsuperscript{\ref{note1}}    &  -   & CLIP(ViT-B-32)   & 151M    & \textcolor{blue}{\textbf{13.54}}                & \textcolor{blue}{\textbf{30.83}}                & \textcolor{blue}{\textbf{43.46}}                & \textcolor{blue}{\textbf{11.55}}                & \textcolor{blue}{\textbf{33.14}}                & \textcolor{blue}{\textbf{49.83}}                & \textcolor{blue}{\textbf{30.39}}                \\ \hline
CoOp \cite{Zhou2022CoOp}   &  \textit{IJCV'22}  &  CLIP(ViT-B-32)  &  0.10M                  &  6.32                             & 15.89                             & 27.60                              & 4.31                              & 17.89                              & 31.73                              & 17.29                              \\
VPT \cite{10.1007}     &  \textit{ECCV'22}     &  CLIP(ViT-B-32)  &  0.46M              & 7.23                              & 19.40                              & 30.77                              & 6.94                              & 25.26                              & 40.73                              & 21.72                              \\
VL-Prompt     &  -        &  CLIP(ViT-B-32)  &  0.47M            & 6.47                              & 17.05                              & 28.68                              & 6.69                              & 22.62                              & 36.16                              & 19.61                              \\ \hline
Adapter \cite{houlsby19a}   &  \textit{ICML'19}        &  CLIP(ViT-B-32)  &  0.17M          & 8.73                              & 24.73                              & 37.81                              & 8.43                              & 26.02                              & 43.33                             & 24.84                              \\
CLIP-Adapter \cite{10.48550}  &  \textit{arXiv'21}  &  CLIP(ViT-B-32)  & 0.52M            & 7.11                          & 19.48                         & 31.01                         & 7.67                          & 24.87                         & 39.73                         & 21.65                         \\
Convpass \cite{jie2022convolutional}  &  \textit{arXiv'22}  &  CLIP(ViT-B-32)  & 0.14M                 & 6.54                             & 19.67                             & 32.78                              & 7.03                              & 23.08                              & 39.15                              & 21.38                              \\
AdaptFormer \cite{chen2022adaptformer}   &   \textit{NIPS'22}   &   CLIP(ViT-B-32) &  0.17M         & 12.46                              & 28.49                              & 41.86                              & 9.09                              & 29.89                              & 46.81                              & 28.10                             \\  
 Cross-Modal Adapter \cite{Jiang10.48550}    &  \textit{arXiv'22}  &   CLIP(ViT-B-32) & 0.16M            & 11.18                             & 27.31                              & 40.62                              & 9.57                              & 30.74                             & 48.36                              & 27.96                             \\
UniAdapter \cite{2302.06605}    & \textit{arXiv'23}   &  CLIP(ViT-B-32)  & 0.55M          & 12.65                             & 30.81                              & 42.74                              & 9.61                              & 30.06                             & 47.16                              & 28.84                             \\  
\rowcolor{blue!10} Ours &  -  & CLIP(ViT-B-32) & 0.16M  & \textcolor{red}{\textbf{14.13}} & \textcolor{red}{\textbf{31.51}} & \textcolor{red}{\textbf{44.78}} & \textcolor{red}{\textbf{11.63}} & \textcolor{red}{\textbf{33.92}} & \textcolor{red}{\textbf{50.73}} & \textcolor{red}{\textbf{31.12}} \\ \hline
\end{tabular}
\label{tab:results_rsicd}
\end{table*}

\begin{table*}[]
\centering
\caption{
The retrieval performance of state-of-the-art methods on RSITMD tese set. The best performance of traditional methods is with bold.
\textcolor{red}{Red} and \textcolor{blue}{blue} represent the best and second performance in CLIP-based and transfer learning methods.
}
\begin{tabular}{lcccccccccc}
\hline
     &   &   &    & \multicolumn{3}{c}{\textbf{Text retrieval}}                                                                       & \multicolumn{3}{c}{\textbf{Image retrieval}}                                                                      &                               \\
\multirow{-2}{*}{\textbf{Methods}}  &  \multirow{-2}{*}{\textbf{Reference}}    &  \multirow{-2}{*}{\textbf{Backbone (image/text)}} &  \multirow{-2}{*}{\textbf{Params}}   & \textbf{R@1}                  & \textbf{R@5}                  & \textbf{R@10}                 & \textbf{R@1}                  & \textbf{R@5}                  & \textbf{R@10}                 & \multirow{-2}{*}{\textbf{mR}} \\ \hline
\multicolumn{11}{c}{\textbf{\textit{\textbf{Traditional methods}}}} \\ \hline
 VSE++ \cite{VSE}       & \textit{BMVC'18}   & ResNet18, GRU  & 15.78M                  & 10.38                         & 27.65                         & 39.60                         & 7.79                          & 24.87                         & 38.67                         & 24.83                         \\
 SCAN t-i \cite{Lee_2018_ECCV}   & \textit{ECCV'18}   &Faster R-CNN, biGRU   & 13.68M                 & 10.18                         & 28.53                         & 38.49                         & 10.10                         & 28.98                         & 43.53                         & 26.64                         \\
 SCAN i-t \cite{Lee_2018_ECCV}   & \textit{ECCV'18}   & Faster R-CNN, biGRU  & 13.68M                 & 11.06                         & 25.88                         & 39.38                         & 9.82                          & 29.38                         & 42.12                         & 26.28                         \\
 CAMP-triplet \cite{Wang_2019_ICCV}   & \textit{ICCV'19}   & Faster R-CNN, biGRU  & 27.03M             & 11.73                         & 26.99                         & 38.05                         & 8.27                          & 27.79                         & 44.34                         & 26.20                         \\
 CAMP-bce \cite{Wang_2019_ICCV}      & \textit{ICCV'19}   &Faster R-CNN, biGRU   & 27.03M              & 9.07                          & 23.01                         & 33.19                         & 5.22                          & 23.32                         & 38.36                         & 22.03                         \\
 MTFN \cite{10.1145/3343031.3350875}    & \textit{ACM MM'19}   & Faster R-CNN, biGRU  & 77.90M                    & 10.40                         & 27.65                         & 36.28                         & 9.96                          & 31.37                         & 45.84                         & 26.92                         \\
 LW-MCR-b \cite{9594840}    & \textit{TGRS'22}    & SqueezeNet, -   &  1.65M                & 9.07                          & 22.79                         & 38.05                         & 6.11                          & 27.74                         & 49.56                         & 25.55                         \\
 LW-MCR-d \cite{9594840}       & \textit{TGRS'22}   & SqueezeNet, -   & 1.65M            & 10.18                         & 28.98                         & 39.82                         & 7.79                          & 30.18                         & 49.78                         & 27.79                         \\
 LW-MCR-u \cite{9594840}    & \textit{TGRS'22}   & SqueezeNet, -   &  1.65M               & 9.73                          & 26.77                         & 37.61                         & 9.25                          & 34.07                         & 54.03                         & 28.58                         \\
 AMFMN-soft \cite{9437331}     & \textit{TGRS'22}   & ResNet18, biGRU  & 35.94M              & 11.06                         & 25.88                         & 39.82                         & 9.82                          & 33.94                         & 51.90                         & 28.74                         \\
 AMFMN-fusion \cite{9437331}     & \textit{TGRS'22}   & ResNet18, biGRU  &35.94M           & 11.06                         & 29.20                         & 38.72                         & 9.96                          & 34.03                         & 52.96                         & 29.32                         \\
 AMFMN-sim \cite{9437331}    & \textit{TGRS'22}   & ResNet18, biGRU  & 35.94M               & 10.63                         & 24.78                         & 41.81                         & \textbf{11.51}                & 34.69                         & \textbf{54.87}                & 29.72                         \\
 MCRN \cite{YUAN2022103071}    & \textit{JAG'22}    & ResNet18, biGRU  & 52.35M             & 13.27                         & 29.42                         & 41.59                & 9.42                        & 35.53                         & 52.74                         & 30.33                         \\
 GaLR w/o MR \cite{9745546}   & \textit{TGRS'22}   & ResNet18, biGRU  &  46.89M             & 13.05                         & 30.09                         & \textbf{42.70}                & 10.47                         & 36.34                         & 53.35                         & 31.00                         \\
 GaLR with MR \cite{9745546}     & \textit{TGRS'22}    & ResNet18, biGRU  & 46.89M       & \textbf{14.82}                & \textbf{31.64}                & 42.48                         & 11.15                         & \textbf{36.68}                & 51.68                         & \textbf{31.41}                \\ \hline 
\multicolumn{11}{c}{\textbf{\textit{\textbf{CLIP-based methods}}}} \\ \hline
 Single Language \cite{9925582}      & \textit{JSTARS'22}   & CLIP(ViT-B-32)  & 151M       & 19.69                         & 40.26                         & 54.42                         & 17.61                         & 49.73                         & 66.59                         & 41.38                         \\
 Zero-shot CLIP \cite{pmlr-v139-radford21a}     & \textit{ICML'21}   & CLIP(ViT-B-32)  & 0.00M          & 9.29                          & 26.33                         & 37.39                         & 7.79                          & 23.67                         & 38.89                         & 23.89                         \\
 Linear probe \cite{pmlr-v139-radford21a}    & \textit{ICML'21}    & CLIP(ViT-B-32)  & 1.05M            & 17.02                         & 33.12                         & 48.35                         & 13.33                         & 41.80                         & 63.89                         & 36.25                         \\
 Full fine-tuning     & -   & CLIP(ViT-B-32)  & 151M        & \textcolor{red}{\textbf{24.16}}                & \textcolor{red}{\textbf{47.12}}                & \textcolor{red}{\textbf{61.28}}                & \textcolor{red}{\textbf{20.40}}               & \textcolor{blue}{\textbf{50.53}}               & \textcolor{red}{\textbf{68.54}}              & \textcolor{red}{\textbf{45.33}}                \\ \cline{1-11}
 CoOp \cite{Zhou2022CoOp}      & \textit{IJCV'22}   & CLIP(ViT-B-32)  & 0.10M                 & 12.19                         & 30.69                         & 42.82                         & 9.16                          & 33.85                         & 54.35                         & 30.51                         \\
 VPT \cite{10.1007}      & \textit{ECCV'22}   & CLIP(ViT-B-32)  & 0.46M                    & 14.98                         & 32.05                        & 40.15                         & 15.97                         & 41.35                         & 60.35                         & 34.14                         \\
 VL-Prompt           & -   & CLIP(ViT-B-32)  & 0.47M                & 12.81                              & 31.28                              & 42.64                              & 12.61                              & 36.20                              & 58.84                              & 32.40                              \\ \cline{1-11}
 Adapter \cite{houlsby19a}        & \textit{ICML'19}   & CLIP(ViT-B-32)  &  0.17M            & 13.75                         & 27.64                         & 39.96                         & 12.89                         & 40.09                         & 59.91                         & 32.37                         \\
 CLIP-Adapter \cite{10.48550}   & \textit{arXiv'21}   & CLIP(ViT-B-32)  & 0.52M             & 12.83                         & 28.84                         & 39.05                         & 13.30                         & 40.20                         & 60.06                         & 32.38                         \\
 Convpass \cite{jie2022convolutional}    & \textit{arXiv'22}   & CLIP(ViT-B-32)  & 0.14M                  & 16.03                             & 30.16                             & 40.26                              & 12.05                              & 38.66                              & 58.11                              & 32.55                              \\
 AdaptFormer \cite{chen2022adaptformer}      &  \textit{NIPS'22}  &CLIP(ViT-B-32)   & 0.17M           & 16.71                         & 30.16                         & 42.91                         & 14.27                         & 41.53                         & 61.46                         & 34.81                         \\ 
Cross-Modal Adapter \cite{Jiang10.48550}     & \textit{arXiv'22}   & CLIP(ViT-B-32)  & 0.16M             & 18.16                             & 36.08                              & 48.72                             & 16.31                              & 44.33                             & 64.75                              & 38.06                            \\
UniAdapter \cite{2302.06605}      & \textit{arXiv'23}   & CLIP(ViT-B-32)  & 0.55M            & 19.86                             & 36.32                              & 51.28                              & 17.54                              & 44.89                             & 65.46                              & 39.23                             \\  
\rowcolor{blue!10} Ours & - & CLIP(ViT-B-32)   & 0.16M  & \textcolor{blue}{\textbf{23.67}} & \textcolor{blue}{\textbf{44.07}} & \textcolor{blue}{\textbf{60.36}} & \textcolor{blue}{\textbf{20.10}} & \textcolor{red}{\textbf{50.63}} & \textcolor{blue}{\textbf{67.97}} & \textcolor{blue}{\textbf{44.47}} \\ \hline
\end{tabular}

\label{tab:results_rsitmd}
\end{table*}

\begin{table*}[]
\centering
\caption{
The retrieval performance of state-of-the-art methods on UCM tese set. The best performance of traditional methods is with bold.
\textcolor{red}{Red} and \textcolor{blue}{blue} represent the best and second performance in CLIP-based and transfer learning methods.
}
\begin{tabular}{lcccccccccc}
\hline
     &   &   &    & \multicolumn{3}{c}{\textbf{Text retrieval}}                                                                       & \multicolumn{3}{c}{\textbf{Image retrieval}}                                                                      &                               \\
\multirow{-2}{*}{\textbf{Methods}}  &  \multirow{-2}{*}{\textbf{Reference}}    &  \multirow{-2}{*}{\textbf{Backbone (image/text)}} &  \multirow{-2}{*}{\textbf{Params}}   & \textbf{R@1}                  & \textbf{R@5}                  & \textbf{R@10}                 & \textbf{R@1}                  & \textbf{R@5}                  & \textbf{R@10}                 & \multirow{-2}{*}{\textbf{mR}} \\ \hline
\multicolumn{11}{c}{\textbf{\textit{\textbf{Traditional methods}}}} \\ \hline

VSE++ \cite{VSE}       & \textit{BMVC'18} & ResNet18, GRU  & 15.78M            & 12.38                         & 44.76                         & 65.71                         & 10.10                         & 31.80                         & 56.85                         & 36.93                         \\
SCAN t-i \cite{Lee_2018_ECCV}     & \textit{ECCV'18} &Faster R-CNN, biGRU   & 13.68M                 & 14.29                         & 45.71                         & 67.62                         & 12.76                         & 50.38                         & 77.24                         & 44.67                         \\
SCAN i-t \cite{Lee_2018_ECCV}    & \textit{ECCV'18} & Faster R-CNN, biGRU  & 13.68M                & 12.85                         & 47.14                         & 69.52                & 12.48                         & 46.86                         & 71.71                         & 43.43                         \\
CAMP-triplet \cite{Wang_2019_ICCV}     & \textit{ICCV'19} & Faster R-CNN, biGRU  & 27.03M             & 10.95                         & 44.29                         & 65.71                         & 9.90                          & 46.19                         & 76.29                         & 42.22                         \\
CAMP-bce \cite{Wang_2019_ICCV}     & \textit{ICCV'19} &Faster R-CNN, biGRU   & 27.03M               & 14.76                         & 46.19                         & 67.62                         & 11.71                         & 47.24                         & 76.00                         & 43.92                         \\
MTFN \cite{10.1145/3343031.3350875}  & \textit{ACM MM'19} & Faster R-CNN, biGRU  &77.90M                        & 10.47                         & 47.62                         & 64.29                         & \textbf{14.19}                & 52.38                         & 78.95                         & 44.65                         \\
LW-MCR-b \cite{9594840}      &\textit{TGRS'22}  &  SqueezeNet, -  &  1.65M               & 12.38                         & 43.81                         & 59.52                         & 12.00                         & 46.38                         & 72.48                         & 41.10                         \\
LW-MCR-d \cite{9594840}      & \textit{TGRS'22} & SqueezeNet, -   &  1.65M               & 15.24                         & \textbf{51.90}                & 62.86                         & 11.90                         & 50.95                         & 75.24                         & 44.68                         \\
LW-MCR-u \cite{9594840}      & \textit{TGRS'22} & SqueezeNet, -   &  1.65M              & \textbf{18.10}                & 47.14                         & 63.81                         & 13.14                         & 50.38                         & 79.52                & 45.35                         \\
AMFMN-soft \cite{9437331}     & \textit{TGRS'22} & ResNet18, biGRU  & 35.94M               & 12.86                         & \textbf{51.90}                & 66.67                         & \textbf{14.19}                & 51.71                         & 78.48                         & 45.97                         \\
AMFMN-fusion \cite{9437331}     &\textit{TGRS'22}  & ResNet18, biGRU  & 35.94M             & 16.67                         & 45.71                         & 68.57                         & 12.86                         & 53.24                & 79.43                         & 46.08                \\
AMFMN-sim \cite{9437331}   &\textit{TGRS'22}  & ResNet18, biGRU  & 35.94M              & 14.76                         & 49.52                         & 68.10                         & 13.43                         & 51.81                         & 76.48                         & 45.68                         \\
CABIR \cite{app122312221}    &\textit{AS'22}  &  ResNet152, BERT+biGRU  & -               & 15.17                         & 45.71                         & \textbf{72.85}                         & 12.67                         & \textbf{54.19}                         & \textbf{89.23}                         & \textbf{48.30}                         \\ \hline
\multicolumn{11}{c}{\textbf{\textit{\textbf{CLIP-based methods}}}} \\ \hline
Single Language \cite{9925582}     & \textit{JSTARS'22} &CLIP(ViT-B-32)   & 151M          & \textcolor{blue}{\textbf{19.04}}              & 53.33                         & 77.61                         & \textcolor{red}{\textbf{19.33}}                & \textcolor{red}{\textbf{64.00}}                & 91.42                         & 54.12                \\
Zero-shot CLIP \cite{pmlr-v139-radford21a}     &\textit{ICML'21}  & CLIP(ViT-B-32)  & 0.00M           & 10.95                         & 34.76                         & 55.71                         & 7.33                          & 35.14                         & 54.57                         & 33.08                         \\
Linear probe \cite{pmlr-v139-radford21a}      &\textit{ICML'21}  & CLIP(ViT-B-32)  & 1.05M            & 13.33                         & 52.71                         & 77.62                         & 14.43                         & 59.42                         & 90.28                         & 51.30                         \\
Full fine-tuning        & - & CLIP(ViT-B-32)  & 151M      & 17.14                         & \textcolor{blue}{\textbf{55.24}}                & 79.52                & 13.90                         & 56.95                         & 91.81               & 52.43                         \\ \cline{1-11}
CoOp \cite{Zhou2022CoOp}     & \textit{IJCV'22} & CLIP(ViT-B-32)  & 0.10M                     & 7.19                              &  42.76                             & 74.19                              & 9.78                              & 48.34                              & 87.41                              & 44.95                              \\
 VPT \cite{10.1007}        & \textit{ECCV'22} & CLIP(ViT-B-32)  & 0.46M                   & 8.86                              & 47.81                              & 78.62                              & 13.64                              & 55.61                              & 94.05                              & 49.77                              \\
VL-Prompt            & - & CLIP(ViT-B-32)  &  0.47M               & 7.97                              & 45.48                              & 75.96                              & 12.51                              & 51.57                              & 91.52                              & 47.50                              \\ \cline{1-11}
Adapter \cite{houlsby19a}       & \textit{ICML'19} &CLIP(ViT-B-32)   & 0.17M                & 10.17                              & 49.73                              & 80.08                              & 18.20                              & 62.47                              & \textcolor{red}{\textbf{95.77}}                              & 52.74                              \\
CLIP-Adapter \cite{10.48550}    &\textit{arXiv'21}  &CLIP(ViT-B-32)   &  0.52M            & 9.83                         & 42.52                         & 66.79                         & 14.61                         & 51.03                         & 84.14                         & 44.82                         \\
Convpass \cite{jie2022convolutional}       & \textit{arXiv'22} & CLIP(ViT-B-32)  & 0.14M                 & 16.46                             & 54.79                             & 78.78                             & 14.24                              & 58.67                              & 94.51                             & 52.91                              \\
AdaptFormer \cite{chen2022adaptformer}     & \textit{NIPS'22} & CLIP(ViT-B-32)  & 0.17M            & 16.92                             & 54.39                              & 77.46                              & 18.74                              & \textcolor{blue}{\textbf{63.49}}                              & 91.16                              & 53.69                            \\  
Cross-Modal Adapter \cite{Jiang10.48550}        & \textit{arXiv'22} & CLIP(ViT-B-32)  & 0.16M            &13.77                             & 50.57                              & {\textcolor{blue}{\textbf{81.41}}}                              & 17.43                              & 61.45                             & 95.48                              & 53.62                             \\
UniAdapter \cite{2302.06605}      & \textit{arXiv'23} &CLIP(ViT-B-32)   & 0.55M             & 14.46                             & 55.19                              & \textcolor{red}{\textbf{83.95}}                             & 16.74                              & 61.43                             & {\textcolor{blue}{\textbf{95.76}}}                               & \textcolor{blue}{\textbf{54.59}}                             \\  
 \rowcolor{blue!10} Ours & - & CLIP(ViT-B-32) & 0.16M  &  \textcolor{red}{\textbf{22.71}} & \textcolor{red}{\textbf{55.81}} & 80.33 & \textcolor{blue}{\textbf{18.82}} & 62.84 & 93.72 & \textcolor{red}{\textbf{55.71}} \\ \hline
\end{tabular}
\label{tab:results_ucm}
\end{table*}

\subsection{Results Analysis}
\label{sec:analysis}

\begin{figure}[t]	
	\centering
	\includegraphics[width=0.95\linewidth]{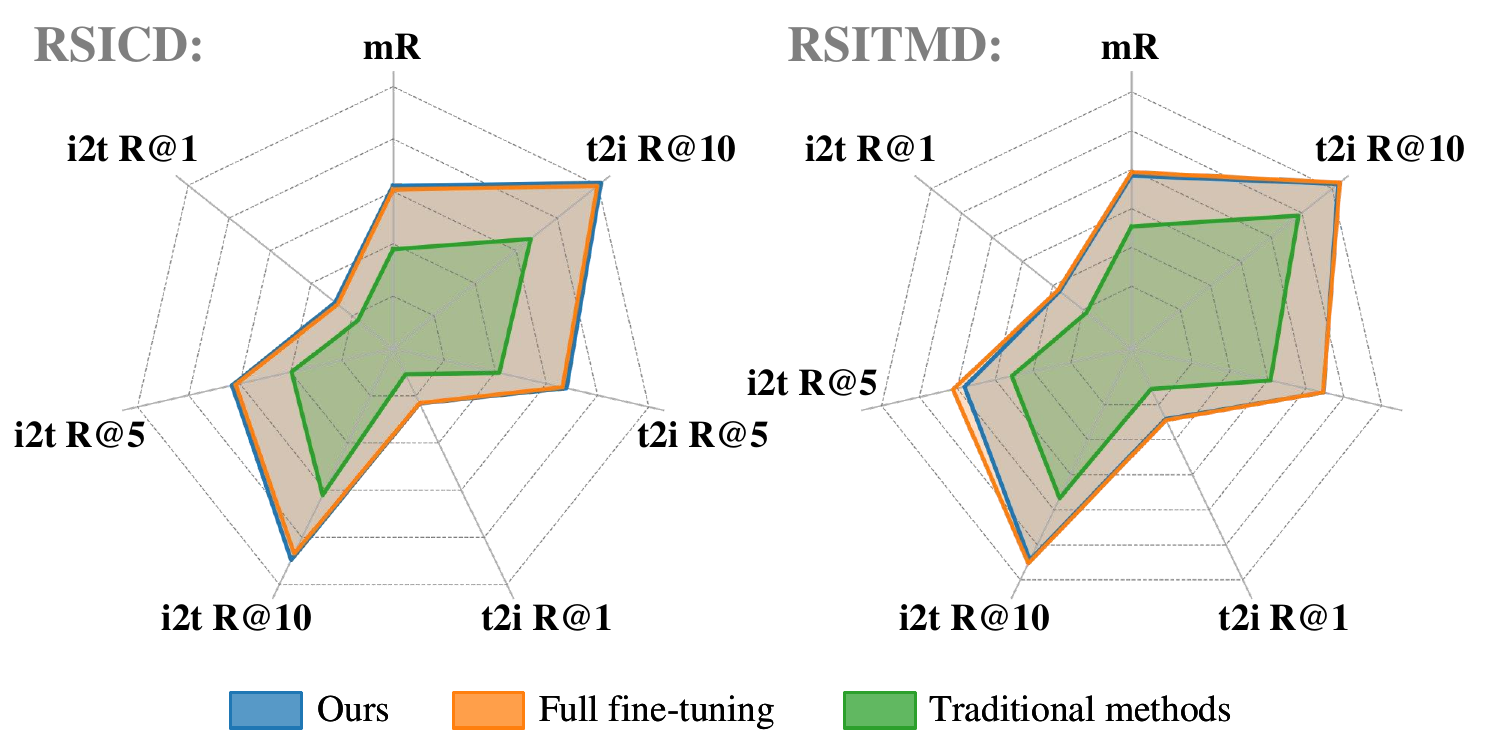}
	\caption{
 Comparison with the RSITR results of full fine-tuning and the best results of traditional methods on RSICD and RSITMD datasets.
 }
	\label{fig:leida}
\end{figure}

\textbf{Results on three datasets.} 
Tables \ref{tab:results_rsicd}, \ref{tab:results_rsitmd}, and \ref{tab:results_ucm} show the test results on the three datasets: RSICD, RSITMD, and UCM. We present results using various traditional methods and CLIP-based fine-tuning methods. Fig. \ref{fig:leida} shows the retrieval results of full fine-tuning and the best results of traditional methods on RSICD and RSITMD datasets. Our method outperforms the traditional methods by 7-13\% and even surpasses the full fine-tuning. CLIP possesses a more complex and larger network structure with a higher number of parameters. 
Compared to traditional methods, PETL, based on CLIP, can achieve significant performance gains by leveraging the powerful generalization ability and rich visual-linguistic prior knowledge obtained through pre-training on massive datasets of natural scenes.

We provide comprehensive empirical studies for PETL-based RS image-text retrieval. 
We observe that our PE-RSITR framework outperforms other works on the RSICD dataset. Compared with full fine-tuning, our results are improved by about 1\%.
Currently, the RSITMD dataset is the highest fine-grained and most challenging RS image-text data. Our approach can obtain optimal and suboptimal results, achieving comparable results to full fine-tuning.
The UCM dataset challenges the robustness of the model due to its small size. As the results are shown in Table \ref{tab:results_ucm}, our method has the most optimal results and exceeds the effect of full fine-tuning. This result demonstrates the robustness of our PE-RSITR framework.

Single Language adopts the dual-transformer structure and loads the pre-trained parameters of CLIP to train. It achieves an average improvement of 8\% in retrieval performance over the best traditional methods on the three datasets.
Zero-shot CLIP, Linear probe, and Full fine-tuning serve as the fundamental baselines for PETL. Zero-shot CLIP directly employs the pre-trained model for testing, resulting in the poorest performance on the three datasets. Full fine-tuning updates all parameters and theoretically yields optimal results. However, due to the limited data size of UCM, full fine-tuning tends to overfit and fails to achieve the best performance, as shown in Table \ref{tab:results_ucm}.

\begin{figure}[t]
	\centering
	\includegraphics[width=0.9\linewidth]{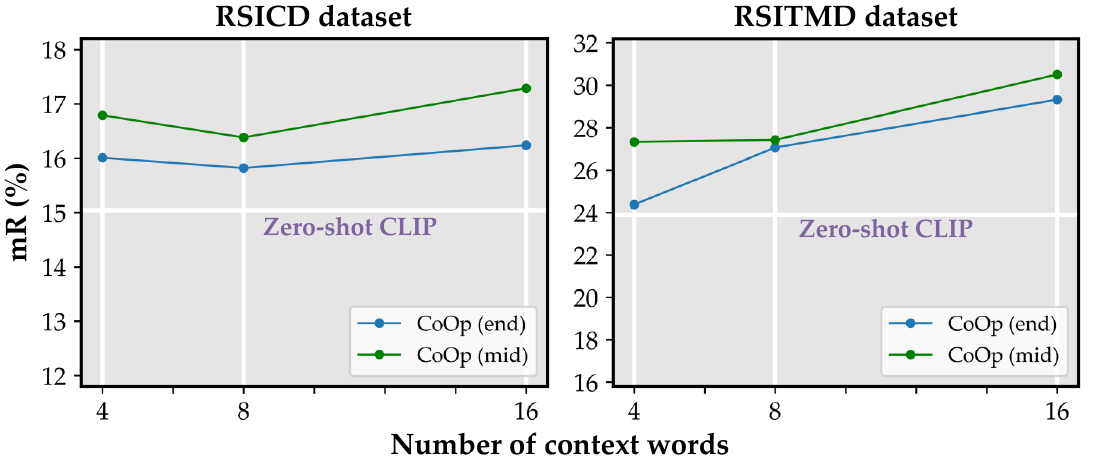}
	\caption{
 Ablation on CoOp's context length. We vary the number of the context length for CoOp-end and CoOp-mid.
 }
	\label{fig:result-coop}
\end{figure}

\begin{figure}[t]
	\centering
	\includegraphics[width=1\linewidth]{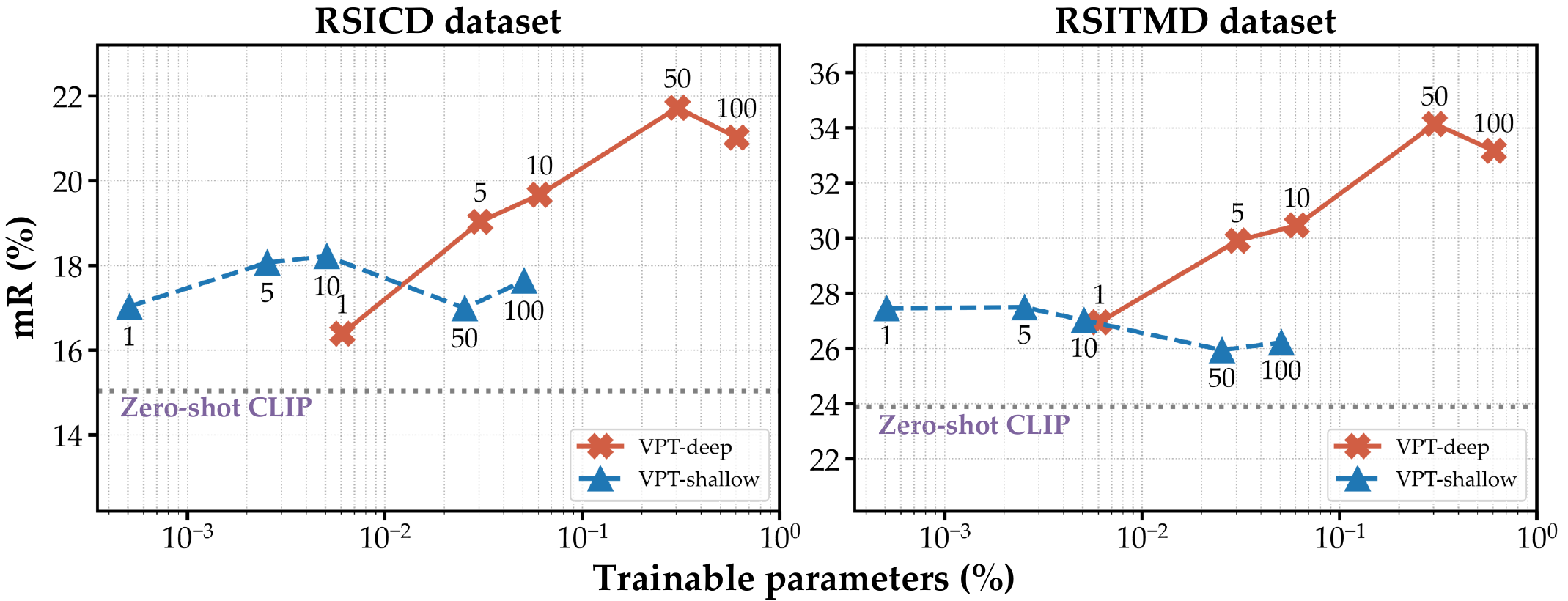}
	\caption{
 Ablation on VPT's prompt length. We vary the number of prompts for VPT-deep and VPT-shallow.
 }
	\label{fig:result-vpt}
\end{figure}

\textbf{Performance analysis on prompt learning.} 
In prompt learning, the VPT achieves superior performance, surpassing traditional methods. The CoOp and VL-Prompt can only achieve performance similar to traditional methods. Due to the significant visual feature gaps between natural and RS domains, adding learnable prompts into visual tokens (VPT) results in a 3-5\% higher retrieval performance compared to text tokens (CoOp). However, the VL-Prompt directly adds prompts to both modalities, without considering multimodal interaction. The increase in prompts makes it more susceptible to overfitting and reduces its generalization ability.

Prompt length is the only additional hyper-parameter needed to tune for prompt learning compared to full fine-tuning. We explore the impact of different prompt lengths on this task and analyze the performance of prompt learning in detail. 
Specifically, we follow the setting of CoOp \cite{Zhou2022CoOp} to vary the context length from 4 to 8 to 16 on RSICD and RSITMD datasets. “end” or “mid” means putting the learnable token in the end or middle. The results are shown in Fig. \ref{fig:result-coop}, which shows that the learnable token in the middle position is better than the end, and more context tokens lead to better performance. However, too many tokens may lead to overfitting.
Likewise, we follow the setting of VPT \cite{10.1007} to vary the prompt length from 1 to 100 to explore two variants: VPT-deep and VPT-shallow. 
As shown in Fig. \ref{fig:result-vpt}, VPT-deep significantly outperforms VPT-shallow. The prompt length has a large impact on the VPT-deep results and shows that the optimal prompt length is 50 for different datasets. More prompts will cause overfitting and more resource consumption.

\textbf{Performance analysis on adapter.} 
Observing the results in Tables \ref{tab:results_rsicd}, \ref{tab:results_rsitmd}, and \ref{tab:results_ucm}, we analyze the performance of adapter tuning. 
CLIP-Adapter \cite{10.48550} only adds a bottleneck layer at the end of the CLIP, making the transfer to the RS domain insufficient. The Adapter \cite{houlsby19a} for language models can only tune textual features alone on the CLIP, and the methods \cite{jie2022convolutional,chen2022adaptformer} for vision models can only tune visual features alone. Using the adapter to adapt the representation in a single branch (image or text) of CLIP is sub-optimal because it does not allow the flexibility to dynamically adapt both representation spaces on the RSITR task. 
UniAdapter and Cross-Modal Adapter are multimodal adapters that support multimodal interaction. However, their structures are complex and fail to consider the problem of high intra-class and inter-class similarity in RS data.
The Cross-Modal Adapter \cite{Jiang10.48550} for multimodal tasks converges faster than UniAdapter \cite{2302.06605}, but the performance of UniAdapter is 1-2\% higher.

The retrieval performance of adapter tuning is 5.9\%, 2.4\%, and 4.7\% higher than that of prompt learning on the three datasets. Prompt learning is an input-related method and adapter tuning is a network-related method. Due to the significant domain gap between the pre-training CLIP and the RS domain, achieving efficient parameter transfer through adding learnable prompts to the input is limited. To adapt to the different knowledge structures of RS image-text and natural image-text, the adapter can adjust the network structure properly to align the RS image and text modality to obtain significant performance gain.

\begin{figure}[t]	
	\centering
	\includegraphics[width=0.95\linewidth]{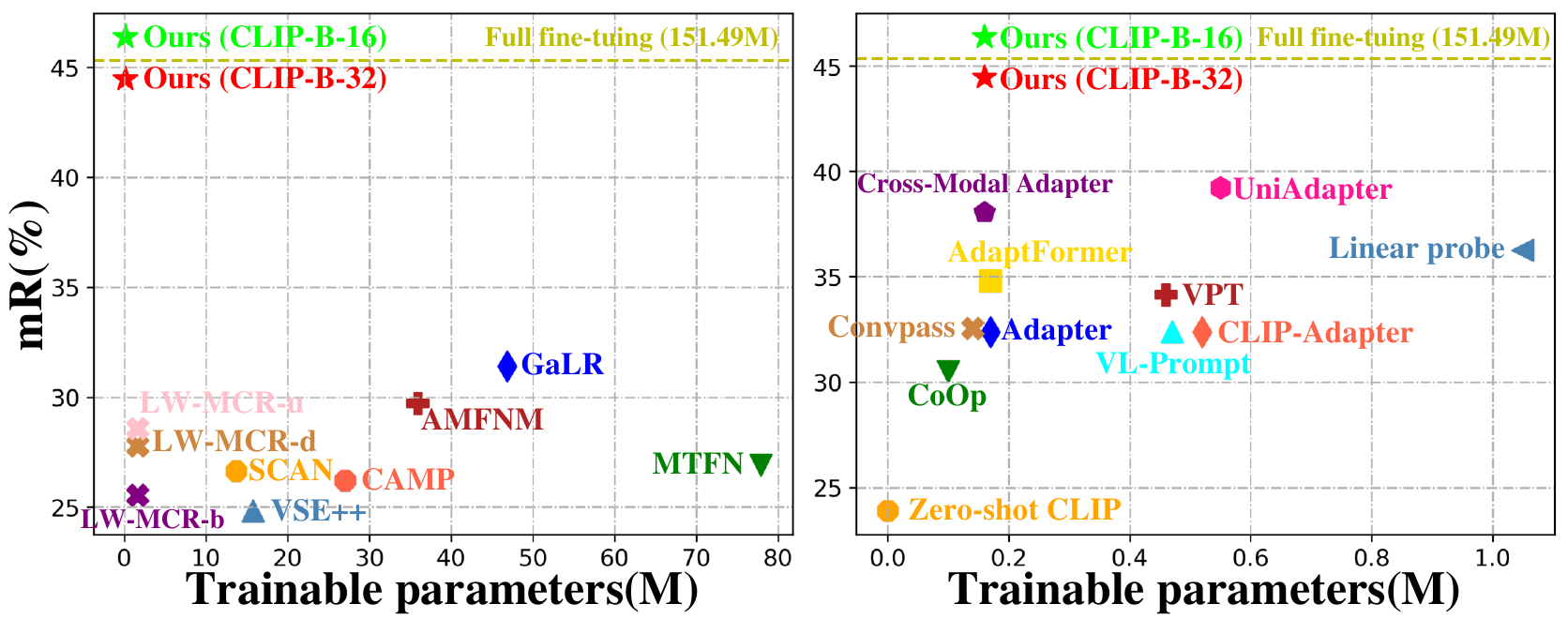}
	\caption{
 Retrieval performance vs. the number of trainable parameters on the RSITMD dataset. \textit{Left}: Comparison with traditional retrieval methods. \textit{Right}: Comparison with CLIP-based methods.
 }
	\label{fig:com-PETL}
\end{figure}

\textbf{Trained parameter efficiency.} 
As shown in Fig. \ref{fig:com-PETL}(a), our method requires the smallest trainable parameters compared to the traditional methods. Even the lightweight model LW-MCR \cite{9594840} requires 1.65 M parameters, our method requires only 0.16 M to achieve SOTA performance. This suggests that finetuning VL pre-training models in the natural domain have great potential for tasks in the RS domain. For the RS image-text retrieval task, our proposed PETL framework is both parameter-efficient and effective.
Among the various PETL methods in Fig.\ref{fig:com-PETL}(b), the trainable parameters of our method are not the smallest, but it achieves the optimal performance with 98.9\% reduction of the full fine-tuning parameters. Our method (CLIP-B-32) achieves nearly the performance of full fine-tuning on the RSITMD dataset and CLIP-B-16 exceeds it. This shows that it is essential to design the adapter and the PETL framework based on the high intra-modal similarity in RS data, which can achieve better performance.

\subsection{Ablation Study}
\label{sec:ablation}
We conduct detailed ablation experiments to validate the effectiveness of the PE-RSITR framework. We systematically analyze the influence of the CLIP backbone network, the proposed MRS-Adapter, and the HMMC loss function on the experiment results.
In the following subsections, we conduct experiments mainly on the RSITMD dataset. The performance of T2I retrieval is measured by the sum of $R@K$ ($K$=1, 5, and 10) of text retrieval and I2T retrieval is measured by the sum of $R@K$ ($K$=1, 5, and 10) of image retrieval.

\begin{figure}[t]	
	\centering
	\includegraphics[width=0.65\linewidth]{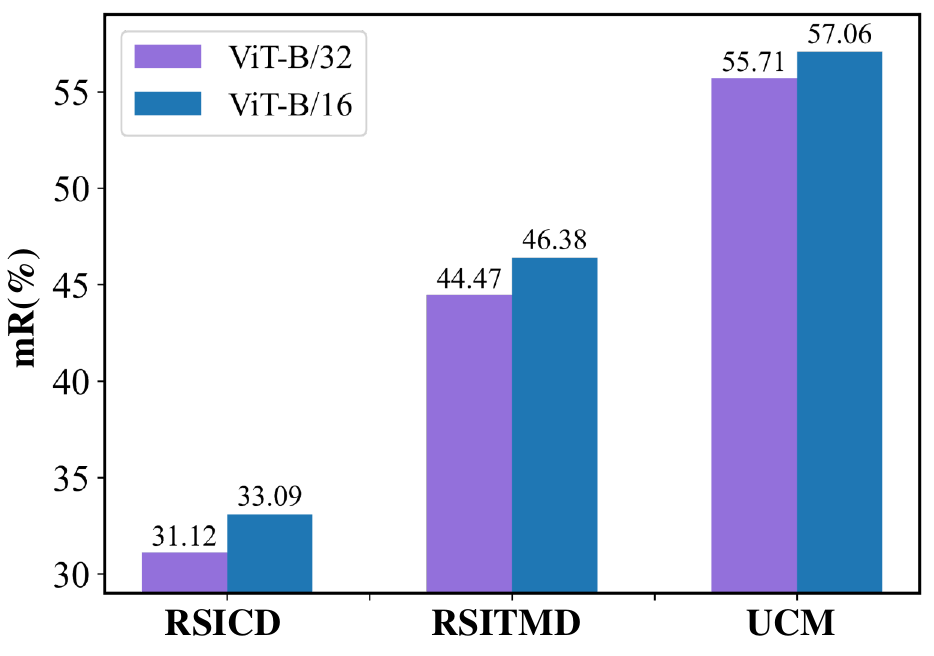}
	\caption{
 Investigations with ViT-B/32 and ViT-B/16 backbones of CLIP.
 }
	\label{fig:visual_backbone}
\end{figure}

\textbf{Vision backbones.}
Fig. \ref{fig:visual_backbone} shows the results on the three datasets for the two ViT backbones of CLIP, \textit{i.e.}, ViT-B/32 and ViT-B/16. The results are expected: the more advanced the backbone, the better the performance.

\textbf{MRS-Adapter.}
Since MRS-Adapter involves dual-modal branches and cross-modal interaction, it is necessary to verify that multimodal branches fine-tuned together are more suitable for the task. We applied it to the visual or textual branch alone or to both branches without the sharing mechanism as comparison experiments.
As shown in Table \ref{tab:branch}, the retrieval performance of applying MRS-Adapter to a single branch is particularly poor, because it cannot adapt the space of the RS domain on both modalities at the same time. This result proves that MRS-Adapter should be used in both visual and text branches. When MRS-Adapter is applied to both branches, the absence of the sharing mechanism can degrade performance.
The weight-sharing benefits the realignment of the VL feature space of CLIP in the RSITR task and can reduce the trained parameters. Specifically, MRS-Adapter reduces the parameters of 0.1M. In summary, MRS-Adapter helps to release the power of pre-trained CLIP models and improve RSITR performance.

\begin{table}[]
    \centering
    \caption{
    Ablations of MRS-Adapter of visual and textual branches.
    }
    \begin{tabular}{ccc|ccc}
    \hline
    Method & Visual & Textual & T2I retrieval & I2T retrieval & mR \\ \hline
    w/o Share    &   & \checkmark       & 107.58        & 120.25        & 37.97   \\
    w/o Share & \checkmark      &       & 114.51        & 130.32        & 40.81   \\
    w/o Share & \checkmark      & \checkmark      &116.47      &136.16        &42.11   \\
    Ours & \checkmark      & \checkmark       & \textbf{128.10}        & \textbf{138.70}        & \textbf{44.47}   \\ \hline
    \end{tabular}
    \label{tab:branch}
\end{table}

\begin{figure}[t]	
	\centering
	\includegraphics[width=0.98\linewidth]{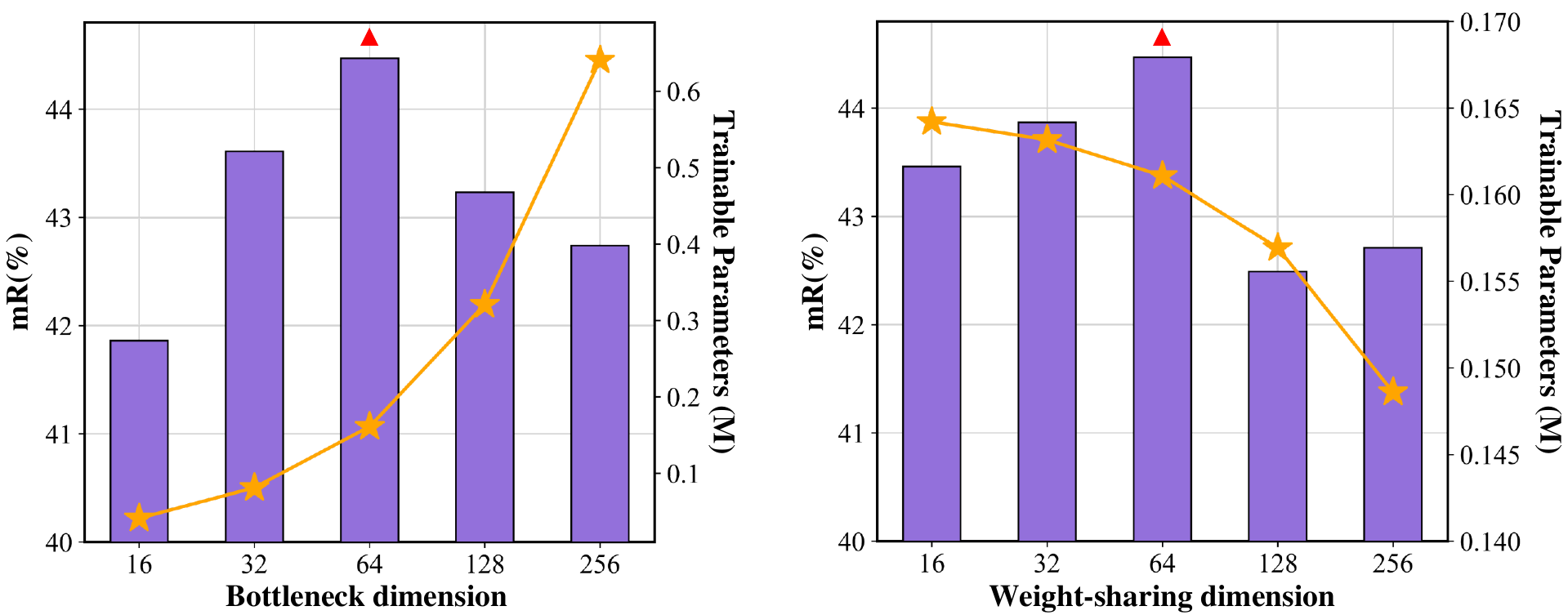}
	\caption{
 \textit{Left}: Ablations of the bottleneck dimension. \textit{Right}: Ablations of the weight-sharing dimension.
 }
	\label{fig:dimension}
\end{figure}

\textbf{Bottleneck dimension $d$.} 
The bottleneck dimension $d$ is an important parameter of MRS-Adapter. We conduct experiments with different sizes of bottleneck, as shown in Fig. \ref{fig:dimension}(left).
As the $d$ increases, the parameters of MRS-Adapter increase. The retrieval performance is poor when the $d$ is less than 64.
It is possible that down-projection loses too much information by projecting features to the lower dimensional space. 
The best performance is achieved when the $d$ is 64, and the performance starts to decrease as the dimension increases further.
Therefore the $d$ of MRS-Adapter is set to 64.

\textbf{Weight-sharing dimension $r$.}
After the bottleneck dimension $d$ is set to 64, we further explore the impact of the weight-sharing dimension $r$. We conduct experiments with different sizes of shared weights, as shown in Fig. \ref{fig:dimension}(right).
The overall effect of the weight-sharing dimension on the results is small. The poor retrieval performance when the $r$ is small may be due to less knowledge sharing between RS image modality and text modality, which cannot fully facilitate modal alignment. The best performance is achieved when the $r$ is 64, and the performance starts to decrease as the dimension increases further. Therefore the $r$ of MRS-Adapter is set to 64.

\textbf{Margins $\lambda$ and $\alpha$.} 
In our HMMC learning objective, the margin parameters serve as crucial hyper-parameters. To observe the influence of parameters $\lambda$, $\alpha_{v}$, and $\alpha_{t}$ on retrieval results, we set up a series of experiments using a controlled variable approach. Regarding the intra-modal constraint, we set $\alpha_{v}$ equal to $\alpha_{t}$. As shown in Table \ref{tab:loss_margin}, the results indicate that the model achieves the best retrieval performance, as measured by mR, when the margin parameters $\lambda$ and $\alpha$ are set to 0.2.

\begin{table}[]
    \centering
    \caption{
    experiment of the HMMC loss with different margins $\lambda$ and $\alpha$.
    }
    \begin{tabular}{cc|ccc}
    \hline
     $\lambda$ & $\alpha$=$\alpha_{v}$=$\alpha_{t}$  &  T2I retrieval & I2T retrieval & mR \\ \hline
     0.2  &    0.1  & 119.91        & 137.94        & 42.97  \\
     0.2  &    0.2  & \textbf{128.10}   & \textbf{138.70}   &\textbf{44.47}  \\
     0.2  &    0.3  & 121.55        & 136.57        & 43.02  \\
     0.2  &    0.4  & 121.08        & 134.48        & 42.51  \\ \hline
     0.1  & 0.2     & 119.68        & 135.73        & 42.57  \\
     0.2  & 0.2     & \textbf{128.10}   & \textbf{138.70}   &\textbf{44.47}  \\
     0.3  & 0.2     & 119.86        & 136.92        & 42.80  \\
     0.4  & 0.2     & 118.72        & 136.28        & 42.50  \\ \hline
    \end{tabular}
    \label{tab:loss_margin}
\end{table}

\textbf{HMMC loss.} 
We have carried out experiments on the hybrid multi-modal contrastive loss, \textit{i.e.}, the inter-modal and intra-modal cooperative constraint framework. We set up four sets of comparisons to observe the effect of increasing the intra-modal constraints on the results, as shown in Table \ref{tab:branch_drop}.
The result shows that the retrieval performance is significantly improved after adding the intra-modal constraints. 
However, when the intra-modal constraint is added to the text branch only, text-to-image retrieval becomes worse, while image retrieval has the best result. Likewise, when the intra-modal constraint is added to the visual modality only, image-to-text retrieval becomes worse, while text retrieval has the best result. The RSITR task requires both text retrieval and image retrieval. Only by adding two intra-modal constraints can achieve a balanced result, and the overall retrieval performance is optimal. This result proves that the proposed HMMC loss is effective in further improving the performance on top of the proposed MRS-Adapter.

\begin{table}[]
    \centering
    \caption{
    Ablations of the hybrid multi-modal contrastive loss.
    }
    \begin{tabular}{cc|ccc}
    \hline
    $\mathcal{L}_{intra-v}$ & $\mathcal{L}_{intra-t}$ & T2I retrieval & I2T retrieval & mR \\ \hline
          &        & 113.29        & 132.14        & 40.91   \\
           & \checkmark       & 121.76        & \textbf{141.02}        & 43.80   \\
    \checkmark      &         & \textbf{128.47}        & 134.81        & 43.88   \\
    \checkmark      & \checkmark       & 128.10        & 138.70        & \textbf{44.47}   \\ \hline
    \end{tabular}
    \label{tab:branch_drop}
\end{table}

\subsection{Qualitative Results}
\label{sec:qualitative}
\begin{figure*}[t]
	\centering
	\includegraphics[width=0.9\linewidth]{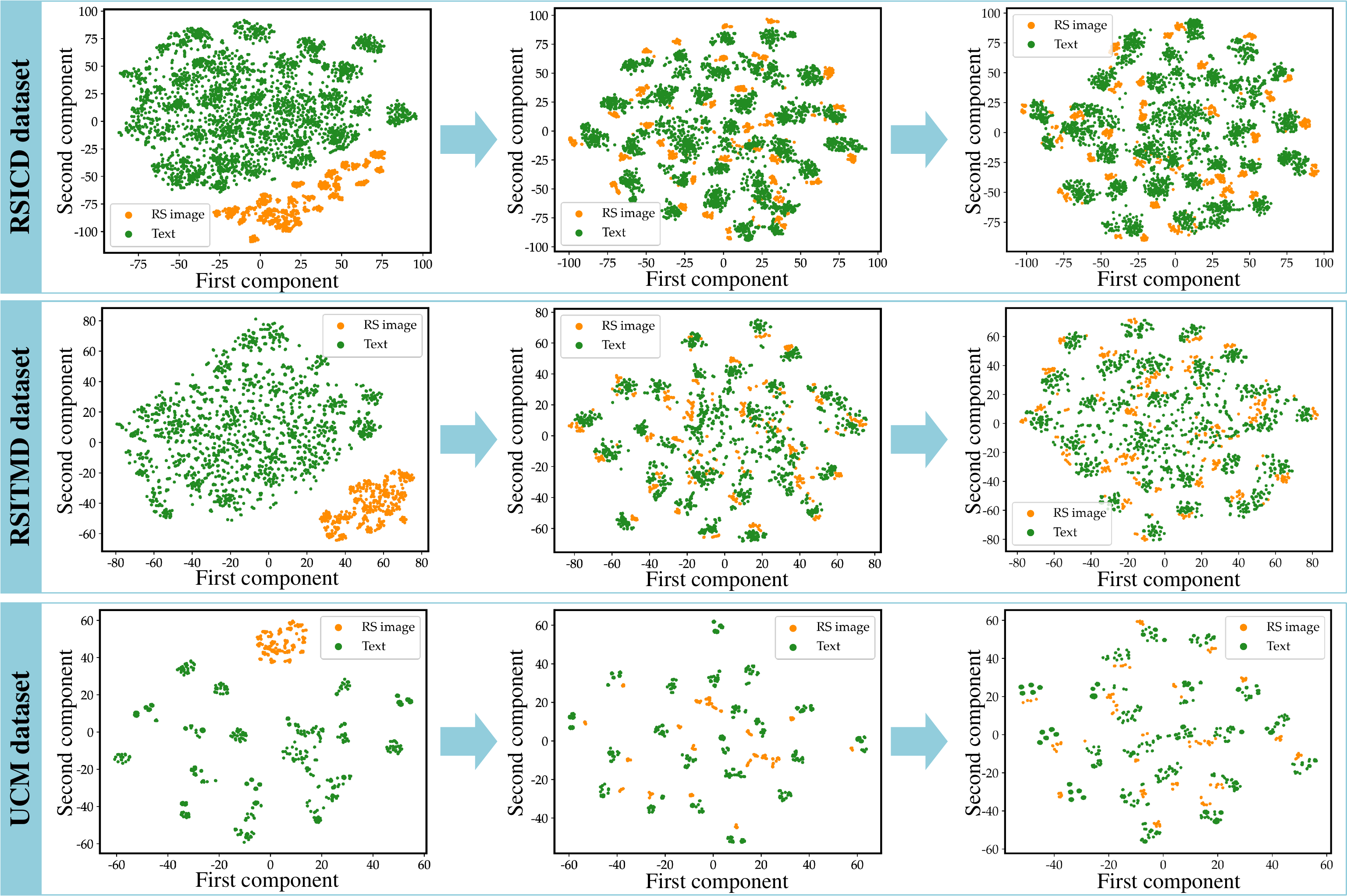}
	\caption{The detailed change of t-SNE visualizations of RS image and text embeddings before and after transfer learning on three datasets. The first column shows the results of original CLIP before transfer learning, the second shows the results after transfer by MRS-Adapter only, and the third shows the results after adding the framework of inter-modal and intra-modal cooperative constraints.
 }
	\label{fig:tSNE}
\end{figure*}

To get an intuition of how the RS image and text are aligned in the joint embedding space, we present a detailed visualization in Fig. \ref{fig:tSNE}. Fig. \ref{fig:tSNE} shows the detailed change of RS image and text embeddings before and after PETL on three datasets. 
We utilize the t-distributed stochastic neighbor embedding (t-SNE) method to project the image and text features obtained from the two encoders into a 2-D space. Observing the first column of Fig. \ref{fig:tSNE}, the CLIP model of the natural domain before transfer learning exhibits a complete separation of RS image and text modalities due to the domain gap.
As shown in the visualization, RSICD has the highest number of sample points. The RSITMD sample points are most scattered and uniformly distributed, indicating the highest fine-grained. The UCM has the lowest number of samples, while different clustering centers indicate different RS scene classes. The unaligned modal representations have semantic inconsistency, thus a large number of errors are sure to occur when Zero-shot CLIP computes the cross-modal similarity.

To tackle this problem, our MRS-Adapter aggregates the embeddings of each modality into a common space, as shown in the second column of Fig. \ref{fig:tSNE}. In addition, it attempts to align image-text pairs separately according to remotely sensed scene-level information to form finer and more discriminative clusters. However, we observe some overlap in each modal representation, which is caused by the great intra-class and inter-class similarity. 
To couple up this problem, we perform the HMMC loss function on top of the proposed MRS-Adapter, as shown in the third column of Fig. \ref{fig:tSNE}. The comparison reveals that our method further spars the distribution of multimodal embeddings. The sample points that originally overlapped in a large number of clusters are now clearly visible. In particular, the results of UCM originally had a lot of text clusters without RS image samples next to them, but now there are RS samples near each cluster of text samples.
The samples of different modalities are more accurately aligned, alleviating the problem of sample overlap and providing a good representation for cross-modal retrieval and matching.

\begin{figure*}
	\centering
	\includegraphics[width=0.95\linewidth]{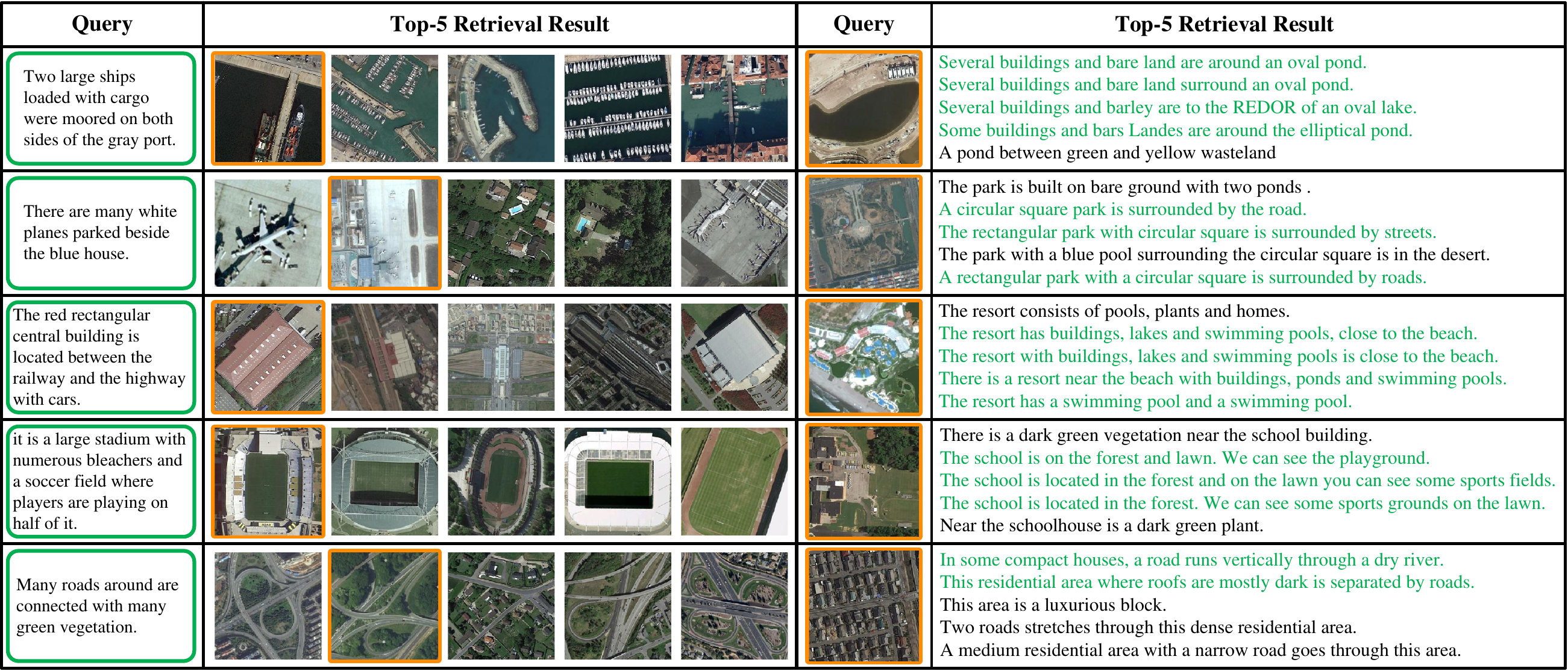}
	\caption{Qualitative top-5 text-to-image and image-to-text retrieval results of our PE-RSITR method on the RSITMD dataset. Images with orange bounding boxes are the ground-truth and the ground-truth texts are marked with green.}
	\label{fig:result}
\end{figure*}

To qualitatively validate the effectiveness of our PE-RSITR method, we displayed several examples of T2I and I2T retrieval in Fig. \ref{fig:result}. Based on these cross-modal retrieval results, our model can accurately distinguish similar samples to retrieve the correct results. Based on these cross-modal retrieval results, our model can accurately distinguish similar samples to retrieve the correct results. The method can understand both abstract phrases and complex long sentences and is robust in the face of both simple and complex images. 
This is mainly attributed to the fact that our designed PE-RSITR framework not only learns the specific knowledge of RS domain, but also exploits the powerful generalization ability and rich VL knowledge structure of CLIP.

\section{Conclusion and Future Work}
\label{sec:conclusion}
In this paper, we explore a new paradigm for the PETL-based RSITR task, namely PE-RSITR, to bridge the differences between the RS domain and the natural domain. Our proposed MRS-Adapter and the HMMC loss function are conceptually simple but can effectively expand CLIP to coarse-grained VL tasks in the RS domain. 
Inserting a lightweight MRS-Adapter into CLIP without modifying the inherent structure consumes only a little computational cost. It can make the best use of the powerful generalization ability and rich visual-language prior knowledge of the CLIP. Constructing intra-modal positive pairs using random dropout and adding intra-modal constraints can effectively avoid the retrieval interference of similar samples. Extensive experiments on RSITR validate that our method can achieve comparable or even better performance than full fine-tuning.

Manual selection of various dimension parameters is required for the MRS-Adapter. How to adaptively select the parameters of the MRS-Adapter remains a challenging topic. Our proposed method does not account for scale variations, cluttered backgrounds, and sparse or dense distribution of objects in RS imagery.
In future work, we will extend our method to other fine-grained perception tasks, such as RS visual grounding (object level) and RS image change captioning (spatiotemporal level). In these tasks, it is essential to extract fine-grained features or similarities from image-text pairs. We hope our work will inspire future research to explore more efficient PETL methods for more fine-grained RS visual-language tasks.


\footnotesize
\bibliographystyle{IEEEtranN}  
\bibliography{document}  

\ifCLASSOPTIONcaptionsoff
  \newpage
\fi

\end{document}